\documentclass[acmlarge]{acmart}
\usepackage{balance}       
\usepackage{graphics}      
\usepackage{hyperref}
\usepackage{color}
\usepackage{booktabs}
\usepackage{textcomp}
\usepackage{subcaption}
\usepackage{enumerate}
\usepackage{xcolor}
\usepackage{lipsum}
\usepackage{makecell}
\usepackage{multicol}
\usepackage{multirow}
\usepackage{array}
\usepackage{verbatimbox}
\usepackage{enumitem}
\usepackage{amsmath}
\usepackage{stfloats}
\usepackage{graphicx}
\usepackage{amsthm}
\usepackage{listings}
\usepackage{caption} 
\usepackage[export]{adjustbox}
\usepackage{xspace}
\usepackage{epsfig}
\usepackage[linesnumbered]{algorithm2e}
\usepackage{algpseudocode}
\usepackage{tabularx}
\usepackage{arydshln}
\usepackage[bottom]{footmisc}
\usepackage{tcolorbox}
\usepackage{stackengine}
\usepackage{placeins}

\newcommand*{\eg}{\textit{e.g.},\xspace}
\newcommand*{\ie}{\textit{i.e.},\xspace}
\newcommand*{\vs}{\textit{vs.}\xspace}
\newcommand*{\etc}{\textit{etc.}}

\newcommand*{\etal}{\textit{et~al.}\xspace}
\newcommand*{\hlinespace}{\addlinespace[1ex]\hline\addlinespace[1ex]}
\newcommand*{\hdashlinespace}{\addlinespace[1ex]\hdashline\addlinespace[1ex]}
\newcommand*{\cdashlinespace}[1]{\addlinespace[1ex]\cdashline{#1}\addlinespace[1ex]}
\newcommand{\clinespace}[1]{\addlinespace[1ex]\cline{#1}\addlinespace[1ex]}

\newcolumntype{L}[1]{>{\raggedright\let\newline\\\arraybackslash\hspace{0pt}}m{#1}}
\newcolumntype{C}[1]{>{\centering\let\newline\\\arraybackslash\hspace{0pt}}m{#1}}
\newcolumntype{R}[1]{>{\raggedleft\let\newline\\\arraybackslash\hspace{0pt}}m{#1}}

\makeatletter
\def\thickhline{%
  \noalign{\ifnum0=`}\fi\hrule \@height \thickarrayrulewidth \futurelet
   \reserved@a\@xthickhline}
\def\@xthickhline{\ifx\reserved@a\thickhline
               \vskip\doublerulesep
               \vskip-\thickarrayrulewidth
             \fi
      \ifnum0=`{\fi}}
\makeatother

\makeatletter
\def\thickhlinespace{%
  \addlinespace[1ex]
  \noalign{\ifnum0=`}\fi\hrule \@height \thickarrayrulewidth \futurelet
   \reserved@a\@xthickhline
   \addlinespace[1ex]
   }
\def\@xthickhlinespace{\ifx\reserved@a\thickhline
               \vskip\doublerulesep
               \vskip-\thickarrayrulewidth
             \fi
      \ifnum0=`{\fi}}
\makeatother

\newlength{\thickarrayrulewidth}
\newlength\Origarrayrulewidth

\algnewcommand{\IfThenElse}[3]{
  \State \algorithmicif\ #1\ \algorithmicthen\ #2\ \algorithmicelse\ #3}
  
\newenvironment{s_itemize}{
\begin{itemize}
  \setlength{\itemsep}{3pt}
  \setlength{\parskip}{0pt}
  \setlength{\parsep}{0pt}
}{\end{itemize}}

\newenvironment{s_enumerate}{
\begin{enumerate}
  \setlength{\itemsep}{3pt}
  \setlength{\parskip}{0pt}
  \setlength{\parsep}{0pt}
}{\end{enumerate}}

\newcommand\downred[1]{\textcolor{downredcolor}{#1}}
\newcommand\upgreen[1]{\textcolor{upgreencolor}{#1}}
\definecolor{downredcolor}{HTML}{e31a1c}
\definecolor{upgreencolor}{HTML}{33a02c}

\definecolor{DarkGreen}{HTML}{5DAC81}
\newcommand\review[1]{\textcolor{black}{#1}}
\newcommand\minorreview[1]{\textcolor{black}{#1}}

\setcopyright{acmlicensed}
\acmJournal{IMWUT}
\acmYear{2024} \acmVolume{8} \acmNumber{1} \acmArticle{32} \acmMonth{3}\acmDOI{10.1145/3643540}

\begin{document}

%

\title{Mental-LLM: Leveraging Large Language Models for Mental Health Prediction via Online Text Data}

%

\author{Xuhai Xu}
\email{xuhaixu@uw.edu}
\orcid{0000-0001-5930-3899}
\affiliation{%
  \institution{Massachusetts Institute of Technology \& University of Washington}
  \country{USA}
}

\author{Bingsheng Yao}
\orcid{0009-0004-8329-4610}
\affiliation{%
  \institution{Rensselaer Polytechnic Institute}
  \country{USA}
}

\author{Yuanzhe Dong}
\orcid{0009-0006-2013-1157}
\affiliation{%
  \institution{Stanford University}
  \country{USA}
}

\author{Saadia Gabriel}
\orcid{0009-0001-9353-951X}
\affiliation{%
    \institution{Massachusetts Institute of Technology}
    \country{USA}
}

\author{Hong Yu}
\orcid{0000-0001-9263-5035}
\affiliation{%
    \institution{University of Massachusetts Lowell}
    \country{USA}
}

\author{James Hendler}
\orcid{0000-0003-3056-1960}
\affiliation{%
  \institution{Rensselaer Polytechnic Institute}
  \country{USA}
}

\author{Marzyeh Ghassemi}
\orcid{0000-0001-6349-7251}
\affiliation{%
    \institution{Massachusetts Institute of Technology}
    \country{USA}
}

\author{Anind K. Dey}
\orcid{0000-0002-3004-0770}
\affiliation{%
  \institution{University of Washington}
  \country{USA}
}
 
\author{Dakuo Wang}
\orcid{0000-0001-9371-9441}
\affiliation{%
  \institution{Northeastern University}
  \country{USA}
}

%
\renewcommand{\shortauthors}{Xu et al.}
\renewcommand{\shorttitle}{Mental-LLM}

%
\begin{abstract}
Advances in large language models (LLMs) have empowered a variety of applications.
However, there is still a significant gap in research when it comes to understanding and enhancing the capabilities of LLMs in the field of mental health.
In this work, we present a comprehensive evaluation of multiple LLMs on various mental health prediction tasks via online text data, including Alpaca, Alpaca-LoRA, FLAN-T5, GPT-3.5, and GPT-4.
We conduct a broad range of experiments, covering zero-shot prompting, few-shot prompting, and instruction fine-tuning. 
The results indicate a promising yet limited performance of LLMs with zero-shot and few-shot prompt designs for mental health tasks.
More importantly, our experiments show that instruction finetuning can significantly boost the performance of LLMs for all tasks simultaneously. 
Our best-finetuned models, Mental-Alpaca and Mental-FLAN-T5, outperform the best prompt design of GPT-3.5 (25 and 15 times bigger) by 10.9\% on balanced accuracy and the best of GPT-4 (250 and 150 times bigger) by 4.8\%. They further perform on par with the state-of-the-art task-specific language model.
We also conduct an exploratory case study on LLMs' capability on mental health reasoning tasks, illustrating the promising capability of certain models such as GPT-4.
We summarize our findings into a set of action guidelines for potential methods to enhance LLMs' capability for mental health tasks.
Meanwhile, we also emphasize the important limitations before achieving deployability in real-world mental health settings, such as known racial and gender bias. We highlight the important ethical risks accompanying this line of research.
\end{abstract}

%
%
%

%
\maketitle

\section{Introduction}
\label{sec:introduction}

The recent surge of Large Language Models (LLMs), such as GPT-4~\cite{bubeck_sparks_2023}, PaLM~\cite{chowdhery_palm_2022}, FLAN-T5~\cite{chung_scaling_2022}, and Alpaca~\cite{taori_stanford_2023}, demonstrates the promising capability of large pre-trained models to solve various tasks in zero-shot settings (\ie tasks not encountered during training). 
Example tasks include question answering~\cite{omar2023chatgpt,robinson2023leveraging}, logic reasoning~\cite{wei_chain--thought_2023,zhou_least--most_2023}, machine translation~\cite{brants2007large,gulcehre2017integrating}, \etc\ 
A number of experiments have revealed that, built on hundreds of billions of parameters, these LLMs have started to show the capability to understand the human common sense beneath the natural language and do proper reasoning and inference accordingly~\cite{bubeck_sparks_2023,nori_capabilities_2023}.

Among different applications, one particular question yet to be answered is how well LLMs can understand human mental health states through natural language.
Mental health problems represent a significant burden for individuals and societies worldwide.
A recent report suggested that more than 20\% of adults in the U.S. experience at least one mental disorder in their lifetime~\cite{mental2022state} and 5.6\% have suffered from a serious psychotic disorder that significantly impairs functioning~\cite{mental2023stats}. The global economy loses around \$1 trillion annually in productivity due to depression and anxiety alone~\cite{mentalcost2023}.

In the past decade, there has been a plethora of research in natural language processing (NLP) and computational social science on detecting mental health issues via online text data such as social media content~(\eg \cite{guntuku_detecting_2017,eichstaedt2018facebook,coppersmith_clpsych_2015,de_choudhury_social_2013,de_choudhury_mental_2014}). However, most of these studies have focused on building domain-specific machine learning (ML) models (\ie one model for one particular task, such as stress detection~\cite{nijhawan2022stress,guntuku2019understanding}, depression prediction~\cite{eichstaedt2018facebook,tadesse2019detection,xu_leveraging_2019, xu_leveraging_2021}, or suicide risk assessment~\cite{de_choudhury_discovering_2016,coppersmith2018natural}). Even for traditional pre-trained language models such as BERT, they need to be finetuned for specific downstream tasks~\cite{devlin_bert_2019,liu_roberta_2019}.
Some studies have also explored the multi-task setup~\cite{benton2017multi}, such as predicting depression and anxiety at the same time~\cite{sarkar2022predicting}. However, these models are constrained to predetermined task sets, offering limited flexibility.
From a different aspect, another line of research has been exploring the application of chatbots for mental health services~\cite{bodrunova_assessing_2019,cameron_towards_2017,lee_designing_2020}. Most chatbots are simply rule-based and can benefit from more advanced models that empower the chatbots~\cite{abd-alrazaq_overview_2019,lee_designing_2020}.
Despite the growing research efforts of empowering AI for mental health, it's important to note that existing techniques can sometimes introduce bias and even provide harmful advice to users~\cite{irene_y_chen_can_2019,timmons2022call,lovejoy2019technology}.

Since natural language is a major component of mental health assessment and treatment~\cite{sharma2018mental,gkotsis2016language}, LLMs could be a powerful tool for understanding end-users' mental states through their written language. 
These instruction-finetuned and general-purpose models can understand a variety of inputs and obviate the need to train multiple models for different tasks. 
Thus, we envision employing a single LLM for a variety of mental health-related tasks, such as multiple question-answering, reasoning, and inference.
This vision opens up a wide range of opportunities for UbiComp, Human-Computer Interaction (HCI), and mental health communities, such as online public health monitoring systems~\cite{patel2018psyheal,graham2019artificial}, \minorreview{mental-health-aware personal chatbots~\cite{abd2021perceptions,denecke2020mental,kernan2023harnessing}, intelligent assistants for mental health therapists~\cite{sharma_towards_2021}, online moderation tools~\cite{franco2023analyzing}, daily mental health counselors and supporters~\cite{sharma_humanai_2023}, \etc\ }
However, there is a lack of investigation into understanding, evaluating, and improving the capability of LLMs for mental-health-related tasks.

There are few recent studies on the evaluation of LLMs (\eg ChatGPT) on mental-health-related tasks, most of which are in zero-shot settings with simple prompt engineering~\cite{yang_evaluations_2023,amin_will_2023,lamichhane_evaluation_2023}. Researchers have shown preliminary results that LLMs have the initial capability of predicting mental health disorders using natural language, with promising but still limited performance compared to state-of-the-art domain-specific NLP models~\cite{yang_evaluations_2023,lamichhane_evaluation_2023}.
This remaining gap is expected since existing general-purpose LLMs are not specifically trained on mental health tasks.
However, to achieve our vision of leveraging LLMs for mental health support and assistance, we need to address the research question: \textbf{How to improve LLMs' capability of mental health tasks}?

We conducted a series of experiments with six LLMs, including Alpaca~\cite{taori_stanford_2023} and 
Alpaca-LoRA (LoRA-finetuned LLaMA on Alpaca dataset)~\cite{hu_lora_2021},
which are representative open-source models focused on dialogue and other tasks; FLAN-T5~\cite{chung_scaling_2022}, a representative open-source model focused on task-solving; \review{LLaMA2~\cite{touvron_llama_2023-1}, one of the most advanced open-source model released by Meta;} GPT-3.5~\cite{noauthor_introducing_2022} and GPT-4~\cite{bubeck_sparks_2023}, representative close-sourced LLMs over 100 billion parameters.
Considering the data availability, we leveraged online social media datasets with high-quality human-generated mental health labels.
Due to the ethical concerns of existing AI research for mental health, we aim to benchmark LLMs' performance as an initial step before moving toward real-life deployment.
Our experiments contained three stages: (1) zero-shot prompting, where we experimented with various prompts related to mental health, (2) few-shot prompting, where we inserted examples into prompt inputs, and (3) instruction-finetuning, where we finetuned LLMs on multiple mental-health datasets with various tasks.

Our results show that the zero-shot approach yields promising yet limited performance on various mental health prediction tasks across all models. Notably, FLAN-T5 and GPT-4 show encouraging performance, approaching the state-of-the-art task-specific model.
Meanwhile, providing a few shots in the prompt can improve the model performance to some extent ($\overline{\Delta}$ = 4.1\%), but the advantage is limited.
Finally and most importantly, we found that instruction-finetuning significantly enhances the model performance across multiple mental-health-related tasks and various datasets simultaneously. 
Our finetuned Alpaca and FLAN-T5, namely \textit{Mental-Alpaca} and \textit{Mental-FLAN-T5}, significantly outperform the best of GPT-3.5 across zero-shot and few-shot settings ($\times$25 and 15 bigger than Alpaca and FLAN-T5) by an average of 10.9\% on balance accuracy, as well as the best of GPT-4 by 4.8\% ($\times$250 and 150 bigger than Alpaca and FLAN-T5). 
Meanwhile, Mental-Alpaca and Mental-FLAN-T5 can further perform on par with the task-specific state-of-the-art Mental-RoBERTa~\cite{ji_mentalbert_2021}. 
We further conduct an exploratory case study on LLM's capability of mental health reasoning (\ie explaining the rationale behind their predictions). Our results illustrate the promising future of certain LLMs like GPT-4, while also suggesting critical failure cases that need future research attention.
We open-source our code and model at \hyperlink{https://github.com/neuhai/Mental-LLM}{https://github.com/neuhai/Mental-LLM}.

\review{Our experiments present a comprehensive evaluation of various techniques to enhance LLMs' capability in the mental health domain.}
However, we also note that our technical results \textit{do not} imply deployability.
There are many important limitations of leveraging LLMs in mental health settings, especially along known racial and gender gaps~\cite{abid2021persistent,ghosh2023chatgpt}.
We discuss the important ethical risks to be addressed before achieving real-world deployment.

\review{
The contribution of our paper can be summarized as follows:
\begin{s_enumerate}
\item We present a comprehensive evaluation of prompt engineering, few-shot, and finetuning techniques on multiple LLMs in the mental health domain.
\item With online social media data, our results reveal that finetuning on a variety of datasets can significantly improve LLM's capability on multiple mental-health-specific tasks across different datasets simultaneously.
\item We release our model \textit{Mental-Alpaca} and \textit{Mental-FLAN-T5} as open-source LLMs targeted at multiple mental health prediction tasks.
\item We provide a few technical guidelines for future researchers and developers on turning LLMs into experts in specific domains. We also highlight the important ethical concerns regarding leveraging LLMs for health-related tasks.
\end{s_enumerate}
}

\section{Background}
\label{sec:background}
We briefly summarize the related work in leveraging online text data for mental health prediction (Sec.~\ref{sub:background:online_text}). We also provide an overview of the ongoing research in LLMs and their application in the health domain (Sec.~\ref{sub:background:LLM}).

\subsection{Online Text Data and Mental Health}
\label{sub:background:online_text}
Online platforms, especially social media platforms, have been acknowledged as a promising lens that is capable of revealing insights into the psychological states, health, and well-being of both individuals and populations~\cite{paul_you_2011,culotta_estimating_2014,chancellor_methods_2020,guntuku_detecting_2017,de_choudhury_mental_2014}.
In the past decade, there has been extensive research about leveraging content analysis and social interaction patterns to identify and predict risks associated with mental health issues, such as anxiety~\cite{ahmed2022machine,saifullah2021comparison,shen_detecting_2017}, major depressive disorder~\cite{park_perception_2021,de_choudhury_predicting_2013,tsugawa_recognizing_2015,de_choudhury_social_2013,xu_leveraging_2021,xu_globem_2022-1}, suicide ideation~\cite{de_choudhury_discovering_2016,ruder_suicide_2011,burnap_machine_2015,tadesse_detection_2019,coppersmith_natural_2018}, and others~\cite{coppersmith_adhd_2015,mitchell_quantifying_2015,coppersmith_measuring_2014,10.1145/3130960}. The real-time nature of social media, along with its archival capabilities, often helps in mitigating retrospective bias. The rich amount of social media data also facilitates the identification, monitoring, and potential prediction of risk factors over time. In addition to observation and detection, social media platforms could further serve as effective channels to offer in-time assistance to communities at risk~\cite{livingston_another_2014,ridout_use_2018,kruzan_social_2022}.

From the computational technology perspective, early research started with basic methods~\cite{coppersmith_measuring_2014,mitchell_quantifying_2015,de_choudhury_social_2013}. For example, pioneering work by Coppersmith \etal~\cite{coppersmith_measuring_2014} employed correlation analysis to reveal the relationship between social media language data and mental health conditions.
Since then, researchers have proposed a wide range of feature engineering methods and built machine-learning models for the prediction~\cite{moreno2011feeling,nguyen2014affective,birnbaum2017collaborative,tsugawa_recognizing_2015,rumshisky2016predicting}. For example, De Choudhury \etal~\cite{de_choudhury_predicting_2013} extracted a number of linguistic styles and other features to build an SVM model to perform depression prediction.
Researchers have also explored deep-learning-based models for mental health prediction to obviate the need for hand-crafted features~\cite{sawhney2018exploring,ji2018supervised}. For instance, Tadesse~\etal~\cite{tadesse_detection_2019} employed an LSTM-CNN model and took word embeddings as the input to detect suicide ideation on Reddit.
More recently, pre-trained language models have become a popular method for NLP tasks, including mental health prediction tasks~\cite{ji_mentalbert_2021,nguyen2022improving,han2022hierarchical}. For example, Jiang~\etal~\cite{jiang2020detection} used the contextual representations from BERT as input features for mental health issue detection.
Otsuka~\etal~\cite{otsukadiagnosing} evaluated the performance of BERT-based pre-trained models in clinical settings.
Meanwhile, researchers have also explored the multi-task setup~\cite{benton2017multi} that aims to predict multiple labels. For example, Sarkar \etal~\cite{sarkar2022predicting} trained a multi-task model to predict depression and anxiety at the same time. However, these multi-task models are constrained to a predetermined task set and thus have limited flexibility.
Our work joins the same goal and aims to achieve a more flexible multi-task capability. We focus on the next-generation technology of instruction-finetuned LLMs, leverage their power in natural language understanding, and explore their capability on mental health tasks with social media data. 

\subsection{LLM and Health Applications}
\label{sub:background:LLM}
After the great success of Transformer-based language models such as BERT~\cite{devlin_bert_2019} and GPT~\cite{radford_improving_2018}, researchers and practitioners have advanced towards larger and more powerful language models (\eg GPT-3~\cite{brown_language_2020} and T5~\cite{raffel_exploring_2020}).
Meanwhile, researchers have proposed instruction finetuning, a method that utilizes varied prompts across multiple datasets and tasks. This technique guides a model during training and generation phases to perform diverse tasks within a single unified framework~\cite{wei_finetuned_2022}.
These instruction-finetuned LLMs, such as GPT-4~\cite{bubeck_sparks_2023}, PaLM~\cite{chowdhery_palm_2022}, FLAN-T5~\cite{chung_scaling_2022}, LLaMA~\cite{touvron_llama_2023}, Alpaca~\cite{taori_stanford_2023}, contain tens to hundreds of billions of parameters, demonstrating promising performance across a variety of tasks, such as question answering~\cite{omar2023chatgpt,robinson2023leveraging}, logic reasoning~\cite{wei_chain--thought_2023,zhou_least--most_2023}, machine translation~\cite{brants2007large,gulcehre2017integrating}, \etc

\minorreview{Researchers have explored the capability of these LLMs in health fields~\cite{jiang_health_2023,singhal_towards_2023,li_chatdoctor_2023,liu_large_2023,wu_pmc-llama_2023,nori_capabilities_2023,lupetti2023trustworthy,}.
For example,
Singhal~\etal~\cite{singhal_towards_2023} finetuned PaLM-2 on medical domains and achieved 86.5\% on MedQA dataset.
Similarly, Wu~\etal~\cite{wu_pmc-llama_2023} finetuned LLaMA on medical papers and showed promising results on multiple biomedical QA datasets.
Jo~\etal~\cite{jo2023understanding} explored the deployment of LLMs for public health scenarios.}
Jiang~\etal~\cite{jiang_health_2023} trained a medical language model on unstructured clinical notes from the electronic health record and fine-tuned it across a wide range of clinical and operational predictive tasks. Their evaluation indicates that such a model can be used for various clinical tasks.

\renewcommand{\arraystretch}{1.3}
\begin{table}[]
\centering
\caption{\review{A Summary of LLM-related Research for Mental Health Applications.}}
\label{tab:related_work}
\resizebox{0.85\textwidth}{!}{
\begin{tabular}{llll}
\thickhlinespace
 & \textbf{LLMs} & \textbf{Methods} & \textbf{Tasks} \\
\thickhlinespace
Lamichhane~\cite{lamichhane_evaluation_2023} & GPT-3.5 & Zero-shot & Classification \\
Amin~\etal~\cite{amin_will_2023} & GPT-3.5 & Zero-shot & Classification \\
Yang~\etal~\cite{yang_evaluations_2023} & GPT-3.5 & Zero-shot & Classification, Reasoning \\
Mental-LLaMA~\cite{yang_mentallama_2023} & LLaMA2, Vicuna (LLaMA-based) & \makecell[l]{Zero-shot, Few-shot,\\Instruction Finetuning} & Classification, Reasoning \\ \hdashline
Mental-LLM (Our Work) & \makecell[l]{Alpaca, Alapca-LoRA, FLAN-T5, \\ LLaMA2, GPT-3.5, GPT-4} & \makecell[l]{Zero-shot, Few-shot,\\Instruction Finetuning} & Classification, Reasoning\\
\thickhlinespace
\end{tabular}
}
\end{table}
\renewcommand{\arraystretch}{1}

There is relatively less work in the mental health domain.
Some work explored the capability of LLMs for sentiment analysis and emotion reasoning~\cite{qin2023chatgpt,zhong2023can,kocon2023chatgpt}.

Closer to our study, Lamichhane~\cite{lamichhane_evaluation_2023} and Amin~\etal~\cite{amin_will_2023} tested the performance of ChatGPT (GPT-3.5) on multiple classification tasks (\eg stress, depression, and suicide detection). The results showed that ChatGPT shows the initial potential for mental health applications, but it still has a great room for improvement, with at least 5-10\% performance gaps on accuracy and F1-score.
Yang~\etal~\cite{yang_evaluations_2023} further evaluated the potential reasoning capability of GPT-3.5 for reasoning tasks (\eg potential stressors).
However, most previous studies focused solely on zero-shot prompting and did not explore other methods to improve the performance of LLMs.
Very recently, Yang~\etal~\cite{yang_mentallama_2023} released Mental-LLaMA, a set of LLaMA-based models finetuned on mental health datasets for a set of mental health tasks.
\minorreview{
Table~\ref{tab:related_work} summarizes the recent related work exploring LLMs' capabilities on mental-health-related tasks.
None of the existing work explores the capability other than LLaMA or GPT-3.5.
In this work, we present a comprehensive and systematic exploration of multiple LLMs' performance on mental health tasks, as well as multiple methods to improve their capabilities.
}

\section{Methods}
\label{sec:methods}
We introduce our experiment design with LMMs on multiple mental health prediction task setups, including zero-shot prompting (Sec.~\ref{sub:methods:zero-shot}), few-shot prompting (Sec.~\ref{sub:methods:few-shot}), and instruction finetuning (Sec.~\ref{sub:methods:finetuning}). These setups are model-agnostic, and we will present the details of language models and datasets employed for our experiment in the next section.

\subsection{Zero-shot Prompting}
\label{sub:methods:zero-shot}
The language understanding and reasoning capability of LLMs have enabled a wide range of applications without the need for any domain-specific data, but only providing appropriate prompts~\cite{kojima_large_2022,wei2021finetuned}.
Therefore, we start with prompt design for mental health tasks in a zero-shot setting.

The goal of prompt design is to empower a pre-trained general-purpose LLM to achieve good performance on tasks in the mental health domain. We propose a general zero-shot prompt template ($\textit{Prompt}_{ZS}$) that consists of four parts:
\begin{equation}
    \textit{Prompt}_{ZS} = \textit{TextData} + \textit{Prompt}_{\textit{Part1-S}} + \textit{Prompt}_{\textit{Part2-Q}} + \textit{OutputConstraint}
\label{eq:prompt-zs}
\end{equation}
where \textit{TextData} is the online text data generated by end-users. \textit{Prompt}$_{\textit{Part1-S}}$ provides specifications for a mental health prediction target. \textit{Prompt}$_{\textit{Part2-Q}}$ poses the question for LLMs to answer. And \textit{OutputConstraint} controls the output of models (\eg ``Only return yes or no'' for a binary classification task).

We propose several design strategies for \textit{Prompt}$_{\textit{Part1-S}}$, as shown in the top part of Table~\ref{tab:prompt_design}: (1) \textbf{Basic}, which leaves it as blank; (2) \textbf{Context Enhancement}, which provides more social media context about the \textit{TextData}; (3) \textbf{Mental Health Enhancement}, which inserts mental health concept by asking the model to act as an expert. (4) \textbf{Context \& Mental Health Enhancement}, which combines both enhancement strategies by asking the model to act as a mental health expert under the social media context.

As for \textit{Prompt}$_{\textit{Part2-Q}}$, we mainly focus on two categories of mental health prediction targets: (1) predicting critical mental states, such as stress or depression, and (2) predicting high-stake risk actions, such as suicide. We tailor the question description for each category. Moreover, for both categories, we explore binary and multi-class classification tasks
\footnote{We also conduct an exploratory case study on mental health reasoning tasks. Please see more details in Sec.~\ref{sub:results:reasoning}.}.
Thus, we also make small modifications based on specific tasks to ensure appropriate questions (see Sec.~\ref{sec:implementation} for our mental health tasks). The bottom part of Table~\ref{tab:prompt_design} summarizes the mapping.

\renewcommand{\arraystretch}{1}
\begin{table}[]
\centering
\caption{Prompt Design for Mental Health Prediction Tasks. \textit{Prompt}$_{\textit{Part1-S}}$ aims to provide a better specification for LLMs and \textit{Prompt}$_{\textit{Part2-Q}}$ poses the questions for LLMs to answer. For Part 1, we propose three strategies: context enhancement, mental health enhancement, and the combination of both. As for Part 2, we design different content for multiple mental health problem categories and prediction tasks. For each part, we propose two to three versions to improve its variation.}
\label{tab:prompt_design}
\label{tab:my-table}
\resizebox{1\textwidth}{!}{
\begin{tabular}{ccl}
\thickhlinespace
 \multicolumn{2}{c}{\textbf{Strategy}} &
  \multicolumn{1}{c}{\textbf{\textit{Prompt}$_{\textit{Part1-S}}$}} \\
\thickhlinespace 
\multicolumn{2}{c}{Basic} &
  \makecell[l]{$\bullet$ \{ blank \} } \\ \hlinespace
\multicolumn{2}{c}{Context Enhancement} &
  \makecell[l]{$\bullet$ This person wrote this paragraph on social media.\\$\bullet$ Consider this post on social media to answer the question.} \\ \hlinespace
\multicolumn{2}{c}{Mental Health Enhancement} &
  \makecell[l]{$\bullet$ As a psychologist, read the post on social media and answer the question.\\ $\bullet$ If you are a psychologist, read the post on social media and answer the question.} \\ \hlinespace
\multicolumn{2}{c}{Context \& Mental Health Enhancement} &
  \makecell[l]{ $\bullet$ This person wrote this paragraph on social media. As a psychologist, read the post on social media \\ \xspace\xspace\xspace\xspace and answer the question.\\ $\bullet$ This person wrote this paragraph on social media. As a psychologist, consider the mental well-being\\\xspace\xspace\xspace\xspace condition expressed in this post, read the post on social media, and answer the question.} \\
\thickhlinespace
\multicolumn{1}{c}{\textbf{Category}} &
  \multicolumn{1}{c}{\textbf{Task}} &
  \multicolumn{1}{c}{\textbf{\textit{Prompt}$_{\textit{Part2-Q}}$}} \\
\thickhlinespace
\multirow{2}{*}{\makecell[c]{\\Mental state\\ (\eg stressed, \\ depressed)}} &
  \makecell[c]{Binary classification\\ (\eg yes or no)} &
  \makecell[l]{$\bullet$ Is the poster {[}stressed{]}?\\$\bullet$ Is the poster of this post {[}stressed{]}?\\$\bullet$ Determine if the poster of this post is {[}stressed{]}.} \\  \clinespace{2-3}
 &
  \makecell[c]{Multi-class classification\\ (\eg multiple levels)} &
  \makecell[l]{$\bullet$ Which level is the person {[}stressed{]}?\\$\bullet$ How {[}stressed{]} is the person?\\ $\bullet$ Determine how {[}stressed{]} the person is.} \\ \hlinespace
\multirow{2}{*}{\makecell[c]{\\Critical risk action\\ (\eg suicide)}} &
  \makecell[c]{Binary classification\\ (\eg yes or no)} &
  \makecell[l]{$\bullet$ Does the poster want to {[}suicide{]}?\\$\bullet$ Is the poster likely to {[}suicide{]}?\\$\bullet$ Determine if the poster of this post want to {[}suicide{]}.} \\  \clinespace{2-3}
 &
  \makecell[c]{Multi-class classification\\ (\eg multiple levels)} &
  \makecell[l]{$\bullet$ Which level of {[}suicide{]} risk does the person have?\\$\bullet$ How {[}suicidal{]} is the person?\\$\bullet$ Determine which level of {[}suicide{]} risk does the person have.}\\
\thickhlinespace
\end{tabular}
}
\end{table}
\renewcommand{\arraystretch}{1.0}

For both \textit{Prompt}$_{\textit{Part1-S}}$ and \textit{Prompt}$_{\textit{Part2-Q}}$, we propose several versions to improve its variability. We then evaluate these prompts on multiple LLMs on different datasets and compare their performance.

\subsection{Few-shot Prompting}
\label{sub:methods:few-shot}

In order to provide more domain-specific information, researchers have also explored few-shot prompting with LLMs by providing few-shot demonstrations to support in-context learning (\eg \cite{agrawal2022large,dang2022prompt}). Note that these few examples are used solely in prompts, and the model parameters remain unchanged. The intuition is to present a few ``examples'' for the model to learn domain-specific knowledge \textit{in situ}.
In our setting, we also test this strategy by adding additional randomly sampled [$\textit{Prompt}_{ZS} - \textit{label}$] pairs. The design of the few-shot prompt ($\textit{Prompt}_{FS}$) is straightforward:
\begin{equation}
    \textit{Prompt}_{FS} = [\textit{Sample Prompt}_{ZS} - \textit{label}]_{M} + \textit{Prompt}_{ZS}
\label{eq:prompt-fs}
\end{equation}
where $M$ is the number of prompt-label pairs and is capped by the input length limit of a model. Note that both the $\textit{Sample Prompt}_{ZS}$ and $\textit{Prompt}_{ZS}$ follow Eq.~\ref{eq:prompt-zs} and employ the same design of \textit{Prompt}$_{\textit{Part1-S}}$ and \textit{Prompt}$_{\textit{Part2-Q}}$ to ensure consistency.

\subsection{Instruction Finetuning}
\label{sub:methods:finetuning}

In contrast to the few-shot prompting strategy in Sec.~\ref{sub:methods:few-shot}, the goal of this strategy is closer to the traditional few-shot transfer learning, where we further train the model with a small amount of domain-specific data (\eg \cite{huang2022large,xu2021raise,liu_large_2023}).
We experiment with multiple finetuning strategies.

\subsubsection{Single-dataset Finetuning}
\label{subsub:methods:finetuning:single-dataset}
Following most of the previous work in the mental health field~\cite{yang_evaluations_2023,coppersmith_clpsych_2015,de_choudhury_discovering_2016}, we first conduct basic finetuning on a single dataset (the training set). This finetuned model can be tested on the same dataset (the test set) to evaluate its performance and different datasets to evaluate its generalizability.

\subsubsection{Multi-dataset Finetuning}
\label{subsub:methods:finetuning:multi-dataset}
From Sec.~\ref{sub:methods:zero-shot} to Sec.~\ref{subsub:methods:finetuning:single-dataset}, we have been focusing on one single mental health dataset $D$.
More interestingly, we further experiment with finetuning across multiple datasets simultaneously. Specifically, we leverage instruction finetuning to enable LLMs to handle multiple tasks in different datasets~\cite{brown_language_2020}.

It is noteworthy that such an instruction finetuning setup differs from the state-of-the-art mental-health-specific models (\eg Mental-RoBERTa~\cite{ji_mentalbert_2021}). The previous models are finetuned for a specific task, such as depression prediction or suicidal ideation prediction. Once trained on task A, the model becomes specific to task A and is only suitable for solving that particular task. In contrast, we finetune LLMs on several mental health datasets, employing diverse instructions for different tasks across these datasets in a single iteration. This enables them to handle multiple tasks without additional task-specific finetuning.


For both single- and multi-dataset finetuning, we follow the same two steps:
\begin{align}
\begin{split}
\text{Step 1:} & \text{ Finetune with } [\textit{Prompt}_{ZS} - \textit{label}]_{\sum\limits^{I} N_{D_{i-train}}} \\
\text{Step 2:} & \text{ Test with } [\textit{Prompt}_{ZS}]_{\sum\limits^{I} N_{D_{i-test}}}
\end{split}
\label{eq:prompt-ft}
\end{align}
where $N_{D_{i-train}}$/$N_{D_{i-test}}$ is the total size of the training/test dataset $D_i$, $I$ represents the set of datasets used for finetuning, and $i$ indicates the specific dataset index ($i \in I, |I| \ge 1$). Both $\textit{Prompt}_{ZS\textit{-train}}$ and $\textit{Prompt}_{ZS\textit{-test}}$ follow Eq.~\ref{eq:prompt-zs}. Similar to the few-shot setup in Eq.~\ref{eq:prompt-fs}, they employ the same design of \textit{Prompt}$_{\textit{Part1-S}}$ and \textit{Prompt}$_{\textit{Part2-Q}}$.


\section{Implementation}
\label{sec:implementation}

Our method design is agnostic to specific datasets or models. In this section, we introduce the specific datasets (Sec.~\ref{sub:implementation:datasets}) and models (Sec.~\ref{sub:implementation:models}) involved in our experiments. In particular, we highlight our instructional-finetuned open-source models Mental-Alpaca and Mental-FLAN-T5 (Sec.~\ref{subsub:implementation:models:mental}).
\review{
We also provide an overview of our experiment setup and evaluation metrics (Sec.~\ref{sub:implementation:setup}).}

\subsection{Datasets and Tasks}
\label{sub:implementation:datasets}

\review{
Our experiment is based on four well-established datasets that are commonly employed for mental health analysis.
These datasets were collected from Reddit due to their high-quality and availability.
It is noteworthy that we intentionally avoid using datasets with weak labels based on specific linguistic patterns (\eg whether a user ever stated ``I was diagnosed with X'').}
Instead, we used ones with human expert annotations or supervision.
We define six diverse mental health prediction tasks based on these datasets.

\begin{s_itemize}
\item \textbf{Dreaddit}~\cite{turcan_dreaddit_2019}: \minorreview{This dataset collected posts via Reddit PRAW API~\cite{Praw-Dev} from Jan 1, 2017 to Nov 19, 2018, which contains ten subreddits in the five domains (abuse, social, anxiety, PTSD, and financial) and includes 2929 users' posts}. Multiple human annotators rated whether sentence segments showed the stress of the poster, and the annotations were aggregated to generate final labels. We used this dataset for a post-level binary stress prediction (\textbf{Task 1}).
\item \textbf{DepSeverity}~\cite{naseem_early_2022}: This dataset leveraged the same posts collected in \cite{turcan_dreaddit_2019}, but with a different focus on depression. Two human annotators followed DSM-5~\cite{regier2013dsm} and categorized posts into four levels of depression: minimal, mild, moderate, and severe. We employed this dataset for two post-level tasks: binary depression prediction (\ie whether a post showed at least mild depression, \textbf{Task 2}), and four-level depression prediction (\textbf{Task 3}).
\item \textbf{SDCNL}~\cite{haque_deep_2021}: \minorreview{This dataset also collected posts from Python Reddit API, including r/SuicideWatch and r/Depressionfrom 1723 users}. Through manual annotation, they labeled whether each post showed suicidal thoughts. We employed this dataset for the post-level binary suicide ideation prediction (\textbf{Task 4}).
\item \textbf{CSSRS-Suicide}~\cite{gaur_knowledge-aware_2019}: \minorreview{This dataset contains posts from 15 mental health-related subreddits from 2181 users between 2005 and 2016}. Four practicing psychiatrists followed Columbia Suicide Severity Rating Scale (C-SSRS) guidelines~\cite{posner2008columbia} to manually annotate 500 users on suicide risks in five levels: supportive, indicator, ideation, behavior, and attempt. We leveraged this dataset for two user-level tasks: binary suicide risk prediction (\ie whether a user showed at least suicide indicator, \textbf{Task 5}), and five-level suicide risk prediction (\textbf{Task 6}).
\end{s_itemize}

In order to test the generalizability of our methods, we also leveraged three other datasets from various platforms. Similarly, all datasets contain human annotations as labels.
\begin{s_itemize}
\item \textbf{Red-Sam}~\cite{kalinathan_data_2022,kayalvizhi2022findings}: \minorreview{This dataset also collected posts with PRAW API~\cite{Praw-Dev} , involving five subreddits (Mental Health, depression, loneliness, stress, anxiety).} Two domain experts' annotations were aggregated to generate depression labels. We used this dataset as an external evaluation dataset on binary depression detection (\textbf{Task 2}). Although also from Reddit, this dataset was not involved in few-shot learning or instruction finetuning. We cross-checked datasets to ensure there were no overlapping posts.
\item \textbf{Twt-60Users}~\cite{jamil_monitoring_2017}: \minorreview{This dataset collected twitters from 60 users during 2015 with Twitter API. Two human annotators labeled every tweet with depression labels}. We used this non-Reddit dataset as an external evaluation dataset on depression detection (\textbf{Task 2}). Note that this dataset has imbalanced labels (90.7\% False), as most tweets did not indicate mental distress.
\item \textbf{SAD}~\cite{mauriello_sad_2021}: This dataset contains SMS-like text messages with nine types of daily stressor categories (work, school, financial problem, emotional turmoil, social relationships, family issues, health, everyday decision-making, and other). \minorreview{These messages were written by 3578 humans.} We used this non-Reddit dataset as an external evaluation dataset on binary stress detection (\textbf{Task 1}). Note that human crowd-workers write the messages under certain stressor-triggered instructions. Therefore, this dataset has imbalanced labels on the other side (94.0\% True).
\end{s_itemize}

\review{Table~\ref{tab:datasets} summarizes the information of the seven datasets and six mental health prediction tasks.}
For each dataset, we conducted an 80\%/20\% train-test split. Notably, to avoid data leakage, each user's data were placed exclusively in either the training or test set.
 
\begin{table}[]
\centering
\caption{\review{Summary of Seven Mental Health Datasets Employed for Our Experiment. The top four datasets are used for both training and testing, while the bottom three datasets are used for external evaluation. We define six diverse mental health prediction tasks on these datasets.}}
\label{tab:datasets}
\resizebox{1\textwidth}{!}{
\begin{tabular}{llll}
\thickhlinespace
\textbf{Dataset} &
  \makecell[c]{\textbf{Task}} &
  \textbf{Dataset Size} &
  \textbf{Text Length (Token)} \\
\thickhlinespace
\makecell[l]{Dreaddit~\cite{turcan_dreaddit_2019}\\\textit{Source: Reddit}} &
  \makecell[l]{\#1: Binary Stress Prediction\\\textit{\xspace\xspace\xspace\xspace post-level}} &
  \makecell[l]{Train: 2838 (47.6\% False, 52.4\% True)\\ Test: 715 (48.4\% False, 51.6\% True)} &
  \makecell[l]{Train: 114 $\pm$ 41\\ Test: 113 $\pm$ 39} \\ \hlinespace
\multirow{7}{*}{\makecell[l]{DepSeverity~\cite{naseem_early_2022}\\\textit{Source: Reddit}}} &
  \makecell[l]{\#2: Binary Depression Prediction\\\textit{\xspace\xspace\xspace\xspace post-level}} &
  \makecell[l]{Train: 2842 (72.9\% False, 17.1\% True)\\ Test: 711 (72.3\% False, 17.7\% True)} &
  \makecell[l]{Train: 114 $\pm$ 41\\ Test: 113 $\pm$ 37} \\ \clinespace{2-4}
 &
  \makecell[l]{\#3: Four-level Depression Prediction\\\textit{\xspace\xspace\xspace\xspace post-level}} &
  \makecell[l]{Train: 2842 (72.9\% Minimum, 8.4\% Mild,\\ \xspace\xspace\xspace\xspace\xspace\xspace\xspace\xspace 11.2\% Moderate, 7.4\% Severe)\\ Test: 711 (72.3\% Minimum, 7.2\% Mild,\\ \xspace\xspace\xspace\xspace\xspace\xspace\xspace\xspace 11.5\% Moderate, 10.0\% Severe)} &
  \makecell[l]{Train: 114 $\pm$ 41\\ \\ Test: 113 $\pm$ 37\\ \xspace} \\ \hlinespace
\makecell[l]{SDCNL~\cite{haque_deep_2021}\\\textit{Source: Reddit}} &
  \makecell[l]{\#4: Binary Suicide Ideation Prediction\\\textit{\xspace\xspace\xspace\xspace post-level}} &
  \makecell[l]{Train: 1516 (48.1\% False, 51.9\% True)\\ Test: 379 (49.1\% False, 50.9\% True)} &
  \makecell[l]{Train: 101 $\pm$ 161\\ Test: 92 $\pm$ 119} \\ \hlinespace
\multirow{4}{*}{\makecell[l]{CSSRS-Suicide~\cite{gaur_knowledge-aware_2019}\\\textit{Source: Reddit}}} &
  \makecell[l]{\#5: Binary Suicide Risk Prediction\\\textit{\xspace\xspace\xspace\xspace user-level}}  &
  \makecell[l]{Train: 400 (20.8\% False, 79.2\% True)\\ Test: 100 (25.0\% False, 75.0\% True)} &
  \makecell[l]{Train: 1751 $\pm$ 2108\\ Test: 1909 $\pm$ 2463} \\ \clinespace{2-4}
 &
  \makecell[l]{\#6: Five-level Suicide Risk Prediction\\\textit{\xspace\xspace\xspace\xspace user-level}} &
  \makecell[l]{Train: 400 (20.8\% Supportive, 20.8\% Indicator,\\ \xspace\xspace\xspace\xspace\xspace\xspace\xspace\xspace 34.0\% Ideation, 14.8\% Behavior, 9.8\% Attempt)\\ Test: 100 (25.0\% Supportive, 16.0\% Indicator,\\ \xspace\xspace\xspace\xspace\xspace\xspace\xspace\xspace 35.0\% Ideation, 18.0\% Behavior, 6.0\% Attempt)} &
  \makecell[l]{Train: 1751 $\pm$ 2108\\ \\ Test: 1909 $\pm$ 2463\\ \xspace }\\
  \thickhlinespace
\makecell[l]{Red-Sam~\cite{kalinathan_data_2022}\\\textit{Source: Reddit}} &
  \makecell[l]{\#2: Binary Depression Prediction\\\textit{\xspace\xspace\xspace\xspace post-level}} &
  \makecell[l]{External Evaluation: 3245 (26.1\% False, 73.9\% True)} &
  \makecell[l]{External Evaluation: 151 $\pm$ 139} \\ \hlinespace
\makecell[l]{Twt-60Users~\cite{jamil_monitoring_2017}\\\textit{Source: Twitter}} &
  \makecell[l]{\#2: Binary Depression Prediction\\\textit{\xspace\xspace\xspace\xspace post-level}} &
  \makecell[l]{External Evaluation: 8135 (90.7\% False, 9.3\% True)} &
  \makecell[l]{External Evaluation: 15 $\pm$ 7} \\ \hlinespace
\makecell[l]{SAD~\cite{mauriello_sad_2021}\\\textit{Source: SMS-like}} &
  \makecell[l]{\#1: Binary Stress Prediction\\\textit{\xspace\xspace\xspace\xspace post-level}} &
  \makecell[l]{External Evaluation: 6185 (6.0\% False, 94.0\% True)} &
  \makecell[l]{External Evaluation: 13 $\pm$ 6} \\
\thickhlinespace
\end{tabular}
}
\end{table}

\subsection{Models}
\label{sub:implementation:models}
We experimented with multiple LLMs with different sizes, pre-training targets, and availability.

\begin{s_itemize}
\item \textbf{Alpaca} (7B)~\cite{taori_stanford_2023}: An open-source large model finetuned from another open-sourced LLaMA 7B model~\cite{touvron_llama_2023} on instruction following demonstrations. Experiments have shown that Alpaca behaves qualitatively similarly to OpenAI’s \texttt{text-davinci-003} on certain task metrics. We choose the relatively small 7B version to facilitate running and finetuning on consumer hardware.
\item \textbf{Alpaca-LoRA} (7B)~\cite{hu_lora_2021}: Another open-source large model finetuned from LLaMA 7B model using the same dataset as Alpaca~\cite{taori_stanford_2023}. This model leverages a different finetuning technique called low-rank adaptation (LoRA)~\cite{hu_lora_2021}, with the goal of reducing finetuning cost by freezing the model weights and injecting trainable rank decomposition matrices into each layer of the Transformer architecture. Despite the similarity in names, it is important to note that Alpaca-LoRA is entirely distinct from Alpaca. They are trained on the same dataset but with different methods.
\item \textbf{FLAN-T5} (11B)~\cite{chung_scaling_2022}: An open-source large model T5~\cite{raffel_exploring_2020} finetuned with a variety of task-based datasets on instructions. Compared to other LLMs, FLAN-T5 focuses more on task solving and is less optimized for natural language or dialogue generation. We picked the largest version of FLAN-T5 (\ie FLAN-T5-XXL), which has a comparable size of Alpaca.
\item \review{\textbf{LLaMA2} (70B)~\cite{touvron_llama_2023-1}: A recent open-source large model released by Meta. We picked the largest version of LLaMA2, whose size is between FLAN-T5 and GPT-3.5.}
\item \textbf{GPT-3.5} (175B)~\cite{noauthor_introducing_2022}: This large model is closed-source and available through API provided by OpenAI. We picked the \texttt{gpt-3.5-turbo}, one of the most capable and cost-effective models in the GPT-3.5 family.
\item \textbf{GPT-4} (1700B)~\cite{bubeck_sparks_2023}: This is the largest closed-source model available through OpenAI API. We picked the \texttt{gpt-4-0613}. Due to the limited availability of API, the cost of finetuning GPT-3.5 or GPT-4 is prohibitive.
\end{s_itemize}

\review{It is worth noting that Alpaca, Alpaca-LoRA, GPT-3.5, LLaMA2 and GPT-4 are all finetuned with natural dialogue as one of the optimization goals.} In contrast, FLAN-T5 is more focused on task-solving.
In our case, the user-written input posts resemble natural dialogue, whereas the mental health prediction tasks are defined as specific classification tasks. It is unclear and thus interesting to explore which LLM fits better with our goal.

\subsubsection{Mental-Alpaca \& Mental-FLAN-T5}
\label{subsub:implementation:models:mental}
Our methods of zero-shot prompting (Sec.~\ref{sub:methods:zero-shot}) and few-shot prompting (Sec.~\ref{sub:methods:few-shot}) do not update model parameters during the experiment.
In contrast, instruction finetuning (Sec.~\ref{sub:methods:finetuning}) will update model parameters and generate new models.
To enhance their capability in the mental health domains, we update Alpaca and FLAN-T5 on six tasks across the four datasets in Sec.~\ref{sub:implementation:datasets} using the multi-dataset instruction finetuning method (Sec.~\ref{subsub:methods:finetuning:multi-dataset}), which leads to our new model \textit{Mental-Alpaca} and \textit{Mental-FLAN-T5}.


\review{
\subsection{Experiment Setup and Metrics}
\label{sub:implementation:setup}
For zero-shot and few-shot prompting methods, we load open-source models (Alpaca, Alpaca-LoRA, FLAN-T5, LLaMA2) with one to eight Nvidia A100 GPUs to do the tasks, depending on the size of the model. For closed-source models (GPT-3.5, and GPT-4), we use OpenAI API to conduct chat completion tasks.
}

\review{
As for finetuning \textit{Mental-Alpaca} and \textit{Mental-FLAN-T5}, we merge the four datasets together and provide instructions for all six tasks (in the training set). We use eight Nvidia A100 GPUs for instruction finetuning. With cross entropy as the loss function, we backpropagate and update model parameters in 3 epochs, with Adam optimizer and a learning rate as 2$e^{-5}$ (cosine scheduler, warmup ratio 0.03).
}

\review{
We focus on balanced accuracy as the main evaluation metric, \ie the mean of sensitivity (true positive rate) and specificity (true negative rate).
We picked this metric since it is more robust to class imbalance compared to the accuracy or F1 score~\cite{brodersen2010balanced,xu_globem_2022-1}.
It is noteworthy that the sizes of LLMs we compare are vastly different, with the number of parameters ranging from 7B to 1700B. A larger model is usually expected to have a better overall performance than a smaller model. We inspect whether this expectation holds in our experiments.
}

\section{Results}
\label{sec:results}
We summarize our experiment results with zero-shot prompting (Sec.~\ref{sub:results:zero-shot}), few-shot prompting (Sec.~\ref{sub:results:few-shot}), and instruction finetuning (Sec.~\ref{sub:results:finetune}).
Moreover, although we mainly focus on prediction tasks in this research, we also present the initial results of our exploratory case study on mental health reasoning tasks in Sec.~\ref{sub:results:reasoning}.

Overall, our results show that zero-shot and few-shot settings show promising performance of LLMs for mental health tasks, although their performance is still limited.
Instruction-finetuning on multiple datasets (Mental-Alpaca and Mental-FLAN-T5) can significantly boost models' performance on all tasks simultaneously.
Our case study also reveals the strong reasoning capability of certain LLMs, especially GPT-4.
However, we note that these results \textit{do not} indicate the deployability. We highlight important ethical concerns and gaps in Sec.~\ref{sec:discussion}.

\subsection{Zero-shot Prompting Shows Promising yet Limited Performance}
\label{sub:results:zero-shot}

\review{We start with the most basic zero-shot prompting with Alpaca, Alpaca-LoRA, FLAN-T5, LLaMA2, GPT-3.5, and GPT-4.}
The balanced accuracy results are summarized in the first sections of Table~\ref{tab:results_overall_short}.
Alpaca$_{ZS}$ and Alpaca-LoRA$_{ZS}$ achieve better overall performance than the naive majority baseline ($\overline{\Delta}_{\text{Alpaca}} = 5.5\%$, $\overline{\Delta}_{\text{Alpaca-LoRA}} = 5.6\%$), but they are far from the task-specific baseline models BERT and Mental-RoBERTa (which have 20\%-25\% advantages).
With much larger models GPT-3.5$_{ZS}$, the performance gets more promising ($\overline{\Delta}_{\text{GPT-3.5}} = 12.4\%$ over baseline), which is inline with previous work~\cite{yang_evaluations_2023}. GPT-3.5's advantage over Alpaca and Alpaca-LoRA is expected due to its larger size (25$\times$).

Surprisingly, FLAN-T5$_{ZS}$ achieves much better overall results compared to Alpaca$_{ZS}$ ($\overline{\Delta}_{\text{FLAN-T5}\_vs\_\text{Alpaca}} = 10.9\%$) and Alpaca-LoRA$_{ZS}$ ($\overline{\Delta}_{\text{FLAN-T5}\_vs\_\text{Alpaca-LoRA}} = 11.0\%$), and even \review{LLaMA2 ($\overline{\Delta}_{\text{FLAN-T5}\_vs\_\text{LLaMA2}} = 1.0\%$) and GPT-3.5 ($\overline{\Delta}_{\text{FLAN-T5}\_vs\_\text{GPT-3.5}} = 4.2\%$). Note that LLaMA2 is 6 times bigger  than FLAN-T5 and GPT-3.5 is 15 times bigger.}
On Task \#6 (Five-level Suicide Risk Prediction), FLAN-T5$_{ZS}$ even outperforms the state-of-the-art Mental-RoBERTa by 4.5\%.
Comparing these results, the task-solving-focused model FLAN-T5 appears to be better at the mental health prediction tasks in a zero-shot setting. We will introduce more interesting findings after finetuning (see Sec.~\ref{subsub:results:finetune:dialogue_vs_tasksolving}).

In contrast, the advantage of GPT-4 becomes relatively less remarkable considering its gigantic size. \mbox{GPT-4$_{ZS}$'s} average performance outperforms FLAN-T5$_{ZS}$ (150$\times$ size), \review{LLaMA2$_{ZS}$ (25$\times$ size)}, and GPT-3.5$_{ZS}$ (10$\times$ size) by 6.4\%, 7.5\%, and 10.6\%, respectively. Yet it is still very encouraging to observe that GPT-4 is approaching the state-of-the-art on these tasks ($\overline{\Delta}_{\text{GPT-4}\_vs\_\text{Mental-RoBERTa}}$ $ = -7.9\%$), and it also outperforms Mental-RoBERTa on Task \#6 by 4.5\%.
In general, these results indicate the promising capability of LLMs on mental health prediction tasks compared to task-specific models, even without any domain-specific information.

\begin{table}[!t]
\centering
\caption{
\review{Balanced Accuracy Performance Summary of Zero-shot, Few-shot and Instruction Finetuning on LLMs.
$ZS_{best}$ highlights the best performance among zero-shot prompt designs, including context enhancement, mental health enhancement, and their combination (see Table.~\ref{tab:prompt_design}). Detailed results can be found in Table~\ref{tab:results_overall} in Appendix.
Small numbers represent standard deviation across different designs of \textit{Prompt}$_{\textit{Part1-S}}$ and \textit{Prompt}$_{\textit{Part2-Q}}$. The baselines at the bottom rows do not have standard deviation as the task-specific output is static, and prompt designs do not apply.
Due to the maximum token size limit, we only conduct few-shot prompting on a subset of datasets and mark other infeasible datasets as ``--''.
For each column, the best result is \textbf{bolded}, and the second best is \underline{underlined}.
}
}
\vspace{-0.3cm}
\label{tab:results_overall_short}
\resizebox{0.95\textwidth}{!}{
\begin{tabular}{llcccccc}
\thickhlinespace
& \makecell[r]{\textbf{Dataset}} & \textbf{Dreaddit} & \multicolumn{2}{c}{\textbf{DepSeverity}} & \textbf{SDCNL} & \multicolumn{2}{c}{\textbf{CSSRS-Suicide}} \\ \addlinespace[1ex]
\textbf{Category} & \textbf{Model}   & \textbf{Task \#1}       & \textbf{Task \#2}             & \textbf{Task \#3}             & \textbf{Task \#4}    & \textbf{Task \#5}     & \textbf{Task \#6} \\ \thickhlinespace
\multirow{16}{*}{\makecell{Zero-shot\\Prompting}} & Alpaca$_{ZS}$         & 0.593$_{\pm0.039}$ & 0.522$_{\pm0.022}$ & 0.431$_{\pm0.050}$ & 0.493$_{\pm0.007}$ & 0.518$_{\pm0.037}$ & 0.232$_{\pm0.076}$ \\
& Alpaca$_{ZS\_best}$ & 0.612$_{\pm0.065}$ & 0.577$_{\pm0.028}$ & 0.454$_{\pm0.143}$ & 0.532$_{\pm0.005}$ & 0.532$_{\pm0.033}$ & 0.250$_{\pm0.060}$ \\ \cdashlinespace{2-8}
& Alpaca-LoRA$_{ZS}$           & 0.571$_{\pm0.043}$ & 0.548$_{\pm0.027}$ & 0.437$_{\pm0.044}$ & 0.502$_{\pm0.011}$ & 0.540$_{\pm0.012}$ & 0.187$_{\pm0.053}$ \\
& Alpaca-LoRA$_{ZS\_best}$   & 0.571$_{\pm0.043}$ & 0.548$_{\pm0.027}$ & 0.437$_{\pm0.044}$ & 0.502$_{\pm0.011}$ & 0.567$_{\pm0.038}$ & 0.224$_{\pm0.049}$ \\ \cdashlinespace{2-8}
 & FLAN-T5$_{ZS}$ & 0.659$_{\pm0.086}$ & 0.664$_{\pm0.011}$ & 0.396$_{\pm0.006}$ & 0.643$_{\pm0.021}$ & 0.667$_{\pm0.023}$ & 0.418$_{\pm0.012}$ \\
 & FLAN-T5$_{ZS\_best}$ & 0.663$_{\pm0.079}$ & 0.674$_{\pm0.014}$ & 0.396$_{\pm0.006}$ & 0.653$_{\pm0.011}$ & 0.667$_{\pm0.023}$ & 0.418$_{\pm0.012}$ \\ \cdashlinespace{2-8}
& LLaMA2$_{ZS}$ & 0.720$_{\pm0.012}$ & 0.693$_{\pm0.034}$  & 0.429$_{\pm0.013}$ & 0.589$_{\pm0.010}$ & 0.691$_{\pm0.014}$ & 0.261$_{\pm0.018}$ \\
& LLaMA2$_{ZS\_best}$ & 0.720$_{\pm0.012}$ & 0.711$_{\pm0.033}$ & 0.444$_{\pm0.021}$ & 0.643$_{\pm0.014}$ & 0.722$_{\pm0.039}$ & 0.367$_{\pm0.043}$ \\ \cdashlinespace{2-8}
& GPT-3.5$_{ZS}$         & 0.685$_{\pm0.024}$ & 0.642$_{\pm0.017}$ & 0.603$_{\pm0.017}$ & 0.460$_{\pm0.163}$ & 0.570$_{\pm0.118}$ & 0.233$_{\pm0.009}$ \\
& GPT-3.5$_{ZS\_best}$ & 0.688$_{\pm0.045}$ & 0.653$_{\pm0.020}$ & 0.642$_{\pm0.034}$ & 0.632$_{\pm0.020}$ & 0.617$_{\pm0.033}$ & 0.310$_{\pm0.015}$ \\ \cdashlinespace{2-8}
 & GPT-4$_{ZS}$ & 0.700$_{\pm0.001}$ & 0.719$_{\pm0.013}$ & 0.588$_{\pm0.010}$ & 0.644$_{\pm0.007}$ & 0.760$_{\pm0.009}$ & 0.418$_{\pm0.009}$ \\
 & GPT-4$_{ZS\_best}$ & 0.725$_{\pm0.009}$ & 0.719$_{\pm0.013}$ & 0.656$_{\pm0.001}$ & 0.647$_{\pm0.014}$ & 0.760$_{\pm0.009}$ & \underline{0.441}$_{\pm0.057}$ \\
\thickhlinespace
\multirow{4}{*}{\makecell{Few-shot\\Prompting}} &  Alpaca$_{FS}$         & 0.632$_{\pm0.030}$ & 0.529$_{\pm0.017}$ & 0.628$_{\pm0.005}$ & ---                & ---                & ---                \\
& FLAN-T5$_{FS}$         & 0.786$_{\pm0.006}$ & 0.678$_{\pm0.009}$ & 0.432$_{\pm0.009}$ & ---                & ---                & ---                \\
& GPT-3.5$_{FS}$         & 0.721$_{\pm0.010}$ & 0.665$_{\pm0.015}$ & 0.580$_{\pm0.002}$ & ---                & ---                & ---                \\
& GPT-4$_{FS}$         & 0.698$_{\pm0.009}$ & 0.724$_{\pm0.005}$ & 0.613$_{\pm0.001}$ & ---                & ---                & ---                \\\thickhlinespace
\multirow{2}{*}{\makecell{Instructional\\Finetuning}} & Mental-Alpaca         & \underline{0.816}$_{\pm0.006}$ & \underline{0.775}$_{\pm0.006}$ & \underline{0.746}$_{\pm0.005}$ & \textbf{0.724}$_{\pm0.004}$ & 0.730$_{\pm0.048}$ & 0.403$_{\pm0.029}$ \\
& Mental-FLAN-T5 &  0.802$_{\pm0.002}$ & 0.759$_{\pm0.003}$ & \textbf{0.756}$_{\pm0.001}$ & 0.677$_{\pm0.005}$ & \textbf{0.868}$_{\pm0.006}$ & \textbf{0.481}$_{\pm0.006}$ \\ \thickhlinespace
\multirow{3}{*}{\xspace\xspace\xspace Baseline} & Majority          & 0.500$_{\pm ---}$  & 0.500$_{\pm ---}$  & 0.250$_{\pm ---}$  & 0.500$_{\pm ---}$  & 0.500$_{\pm ---}$  & 0.200$_{\pm ---}$  \\
& BERT              & 0.783$_{\pm ---}$  & 0.763$_{\pm ---}$  & 0.690$_{\pm ---}$  & 0.678$_{\pm ---}$  & 0.500$_{\pm ---}$  & 0.332$_{\pm ---}$  \\
& Mental-RoBERTa     & \textbf{0.831}$_{\pm ---}$  & \textbf{0.790}$_{\pm ---}$   & 0.736$_{\pm ---}$  & \underline{0.723}$_{\pm ---}$  & \underline{0.853}$_{\pm ---}$  & 0.373$_{\pm ---}$  \\ \thickhlinespace
\end{tabular}
}
\end{table}

\subsubsection{The Effectiveness of Enhancement Strategies.}
In Sec.~\ref{sub:methods:zero-shot}, we propose context enhancement, mental health enhancement, and their combination strategies for zero-shot prompt design to provide more information about the domain. Interestingly, our results suggest varied effectiveness on different LLMs and datasets.

\begin{table}[!t]
\centering
\caption{
\review{
Balanced Accuracy Performance Change using Enhancement Strategies. The green/red color indicates increased/decreased accuracy. This table zooms in on the zero-shot section of Table~\ref{tab:results_overall_short}. \upgreen{$\uparrow$}/\downred{$\downarrow$} marks the ones with better/worse performance in comparison.
}
}
\label{tab:results_zeroshot_delta}
\resizebox{0.93\textwidth}{!}{
\begin{tabular}{lccccccc}
\thickhlinespace
\makecell[r]{\textbf{Dataset}} & \textbf{Dreaddit} & \multicolumn{2}{c}{\textbf{DepSeverity}} & \textbf{SDCNL} & \multicolumn{2}{c}{\textbf{CSSRS-Suicide}} & \\ \addlinespace[1ex]
\textbf{Model}   & \textbf{Task \#1}       & \textbf{Task \#2}             & \textbf{Task \#3}             & \textbf{Task \#4}    & \textbf{Task \#5}     & \textbf{Task \#6} & $\overline{\Delta}-$All Six Tasks \\ \thickhlinespace
$\Delta-$Alpaca$_{ZS\_context}$ & \upgreen{ $\uparrow +0.019$} & \upgreen{ $\uparrow +0.045$} & \upgreen{ $\uparrow +0.023$} & \upgreen{ $\uparrow +0.004$} & \upgreen{ $\uparrow +0.014$} & \upgreen{ $\uparrow +0.018$}  &   \upgreen{ $\uparrow +0.021$} \\
$\Delta-$Alpaca$_{ZS\_mh}$ & \upgreen{ $\uparrow +0.000$} & \upgreen{ $\uparrow +0.055$} & \upgreen{ $\uparrow +0.013$} & \downred{ $\downarrow -0.011$} & \upgreen{ $\uparrow +0.006$} & \upgreen{ $\uparrow +0.004$} &   \upgreen{ $\uparrow +0.011$} \\
$\Delta-$Alpaca$_{ZS\_both}$ & \downred{ $\downarrow -0.053$} & \upgreen{ $\uparrow +0.037$} & \downred{ $\downarrow -0.010$} & \upgreen{ $\uparrow +0.039$} & \downred{ $\downarrow -0.007$} & \downred{ $\downarrow -0.010$} & \downred{ $\downarrow -0.001$} \\ \hdashlinespace
$\Delta-$Alpaca-LoRA$_{ZS\_context}$ & \downred{ $\downarrow -0.035$} & \downred{ $\downarrow -0.047$} & \downred{ $\downarrow -0.094$} & \downred{ $\downarrow -0.030$} & \upgreen{ $\uparrow +0.027$} & \upgreen{ $\uparrow +0.027$} & \downred{ $\downarrow -0.025$} \\
$\Delta-$Alpaca-LoRA$_{ZS\_mh}$ & \downred{ $\downarrow -0.071$} & \downred{ $\downarrow -0.047$} & \downred{ $\downarrow -0.105$} & \downred{ $\downarrow -0.005$} & \upgreen{ $\uparrow +0.017$} & \upgreen{ $\uparrow +0.029$} & \downred{ $\downarrow -0.031$} \\
$\Delta-$Alpaca-LoRA$_{ZS\_both}$ & \downred{ $\downarrow -0.071$} & \downred{ $\downarrow -0.048$} & \downred{ $\downarrow -0.051$} & \downred{ $\downarrow -0.003$} & \downred{ $\downarrow -0.023$} & \upgreen{ $\uparrow +0.037$} & \downred{ $\downarrow -0.027$} \\ \hdashlinespace
$\Delta-$FLAN-T5$_{ZS\_context}$ & \upgreen{$\uparrow$  +0.004} & \upgreen{$\uparrow +0.011$} & \downred{$\downarrow -0.018$} & \upgreen{$\uparrow +0.010$} & \downred{$\downarrow -0.018$} & \downred{$\downarrow -0.040$} & \downred{ $\downarrow -0.009$} \\
$\Delta-$FLAN-T5$_{ZS\_mh}$ & \downred{$\downarrow -0.043$} & \upgreen{$\uparrow +0.003$} & \downred{$\downarrow -0.030$} & \upgreen{$\uparrow +0.005$} & \downred{$\downarrow -0.013$} & \downred{$\downarrow -0.046$} & \downred{ $\downarrow -0.021$} \\
$\Delta-$FLAN-T5$_{ZS\_both}$ & \downred{$\downarrow -0.055$} & \downred{$\downarrow -0.003$} & \downred{$\downarrow -0.007$} & \upgreen{$\uparrow +0.002$} & \downred{$\downarrow -0.010$} & \downred{$\downarrow -0.036$} & \downred{ $\downarrow -0.018$} \\ \hdashlinespace
$\Delta-$LLaMA2$_{ZS\_context}$ & \downred{ $\downarrow -0.062$} &    \upgreen{ $\uparrow +0.014$} &   \downred{ $\downarrow -0.019$} &    \upgreen{ $\uparrow +0.000$} &    \upgreen{ $\uparrow +0.031$} &   \upgreen{ $\uparrow +0.106$} & \upgreen{ $\uparrow +0.012$}\\
$\Delta-$LLaMA2$_{ZS\_mh}$ & \downred{ $\downarrow -0.102$} &    \upgreen{ $\uparrow +0.018$} &   \downred{ $\downarrow -0.033$} &    \upgreen{ $\uparrow +0.053$} &   \upgreen{ $\uparrow +0.004$} &   \upgreen{ $\uparrow +0.031$} & \downred{ $\downarrow -0.005$}\\
$\Delta-$LLaMA2$_{ZS\_both}$ & \downred{ $\downarrow -0.136$} &    \upgreen{ $\uparrow +0.011$} &   \upgreen{ $\uparrow +0.016$} &   \upgreen{ $\uparrow +0.054$} &  \downred{ $\downarrow -0.002$} &    \upgreen{ $\uparrow +0.067$} & \upgreen{ $\uparrow +0.002$} \\ \hdashlinespace
$\Delta-$GPT-3.5$_{ZS\_context}$ & \upgreen{ $\uparrow +0.003$} & \upgreen{ $\uparrow +0.011$} & \downred{ $\downarrow -0.060$} & \upgreen{ $\uparrow +0.157$} & \upgreen{ $\uparrow +0.007$} & \upgreen{ $\uparrow +0.031$} &   \upgreen{ $\uparrow +0.025$} \\
$\Delta-$GPT-3.5$_{ZS\_mh}$ & \downred{ $\downarrow -0.006$} & \downred{ $\downarrow -0.006$} & \upgreen{ $\uparrow +0.039$} & \upgreen{ $\uparrow +0.116$} & \downred{ $\downarrow -0.093$} & \upgreen{ $\uparrow +0.077$} &   \upgreen{ $\uparrow +0.021$} \\
$\Delta-$GPT-3.5$_{ZS\_both}$ & \downred{ $\downarrow -0.005$} & \downred{ $\downarrow -0.015$} & \upgreen{ $\uparrow +0.014$} & \upgreen{ $\uparrow +0.172$} & \upgreen{ $\uparrow +0.047$} & \upgreen{ $\uparrow +0.020$} &   \upgreen{ $\uparrow +0.039$} \\ \hdashlinespace 
$\Delta-$GPT-4$_{ZS\_context}$ & \upgreen{$\uparrow +0.006$} & \upgreen{$\uparrow +0.000$} & \upgreen{$\uparrow +0.001$} & \upgreen{$\uparrow +0.000$} & \downred{$\downarrow -0.007$} & \upgreen{  $\uparrow +0.023$} &   \upgreen{ $\uparrow +0.004$} \\
$\Delta-$GPT-4$_{ZS\_mh}$ & \upgreen{$\uparrow +0.025$} & \downred{$\downarrow -0.035$} & \upgreen{$\uparrow +0.067$} & \upgreen{$\uparrow +0.002$} & \downred{$\downarrow -0.023$} & \downred{$\downarrow -0.022$} &   \upgreen{ $\uparrow +0.002$} \\
$\Delta-$GPT-4$_{ZS\_both}$ & \upgreen{$\uparrow +0.018$} & \downred{$\downarrow -0.031$} & \upgreen{$\uparrow +0.061$} & \upgreen{$\uparrow +0.003$} & \downred{$\downarrow -0.063$} & \downred{$\downarrow -0.006$} & \downred{ $\downarrow -0.003$} \\ \hdashlinespace \\
$\overline{\Delta}-$All Six Models & \downred{$\downarrow -0.031$} & \upgreen{$\uparrow +0.000$} & \downred{$\downarrow -0.011$} & \upgreen{$\uparrow +0.032$} & \downred{$\downarrow -0.006$} & \upgreen{$\uparrow +0.017$} & \upgreen{$\uparrow +0.000$} \\
$\overline{\Delta}-$Alpaca, GPT-3.5, GPT-4 & \upgreen{$\uparrow +0.001$} & \upgreen{$\uparrow +0.007$} & \upgreen{$\uparrow +0.017$} & \upgreen{$\uparrow +0.053$} & \downred{$\downarrow -0.013$} & \upgreen{$\uparrow +0.015$} &  \upgreen{$\uparrow +0.013$}\\
\thickhlinespace
\end{tabular}
}
\end{table}

Table~\ref{tab:results_zeroshot_delta} provides a zoom-in summary of the zero-shot part in Table~\ref{tab:results_overall_short}.
\review{For Alpaca, LLaMA2, GPT-3.5, and GPT-4, the three strategies improved the performance in general ($\overline{\Delta}_{\text{Alpaca}} = 1.0\%$, 13 out of 18 tasks show positive changes; $\overline{\Delta}_{\text{LLaMA2}} = 0.3\%$, 12/18 tasks positive;  $\overline{\Delta}_{\text{GPT-3.5}} = 2.8\%$, 12/18 tasks positive; $\overline{\Delta}_{\text{GPT-4}} = 0.2\%$, 11/18 tasks positive).}
However, for Alpaca-LoRA and FLAN-T5, adding more context or mental health domain information would reduce the model performance ($\overline{\Delta}_{\text{Alpaca-LoRA}} = -2.7\%$, $\overline{\Delta}_{\text{FLAN-T5}} = -1.6\%$). For Alpaca-LoRA, this limitation may stem from being trained with fewer parameters, potentially constraining its ability to understand context or domain specifics. For FLAN-T5, this reduced performance might be attributed to its limited capability in processing additional information, as it is primarily tuned for task-solving.

The effectiveness of strategies on different datasets/tasks also varies. We observe that Task\#4 from the SDCNL dataset and Task\#6 from the CSSRS-Suicide dataset benefit the most from the enhancement.
In particular, GPT-3.5 benefits very significantly from enhancement on Task \#4 ($\overline{\Delta}_{\text{GPT-3.5}-\text{Task\#4}} = 14.8\%$).
\review{And LLaMA2 benefits significantly on Task \#6 ($\overline{\Delta}_{\text{GPT-3.5}-\text{Task\#6}} = 6.8\%$).}
These could be caused by the different nature of datasets. Our results suggest that these enhancement strategies are generally more effective for critical action prediction (\eg suicide, 2/3 tasks positive) than mental state prediction (\eg stress and depression, 1/3 task positive).

\review{We also compare the effectiveness of different strategies on the four models with positive effects: Alpaca, LLaMA2, GPT-3.5, and GPT-4}. The context enhancement strategy has the most stable improvement across all mental health prediction tasks ($\overline{\Delta}_{\text{Alpaca}-context} = 2.1\%$, 6/6 tasks positive; \review{$\overline{\Delta}_{\text{LLaMA2}-context} = 1.2\%$, 4/6 tasks positive;} $\overline{\Delta}_{\text{GPT-3.5}-context} = 2.5\%$, 5/6 tasks positive; $\overline{\Delta}_{\text{GPT-4}-context} = 0.4\%$, 5/6 tasks positive).
Comparatively, the mental health enhancement strategy is less effective ($\overline{\Delta}_{\text{Alpaca}-mh} = 1.1\%$, 5/6 tasks positive; \review{$\overline{\Delta}_{\text{LLaMA2}-mh} = -0.5\%$, 4/6 tasks positive;} $\overline{\Delta}_{\text{GPT-3.5}-mh} = 2.1\%$, 3/6 tasks positive; $\overline{\Delta}_{\text{GPT-4}-mh} = 0.2\%$, 3/6 tasks positive).
The combination of the two strategies yields diverse results. It has the most significant improvement on GPT-3.5's performance, but not on all tasks ($\overline{\Delta}_{\text{GPT-3.5}-both} = 3.9\%$, 4/6 tasks positive), \review{followed by LLaMA2 ($\overline{\Delta}_{\text{LLaMA2}-both} = 0.2\%$, 4/6 tasks positive)}. However, it has slightly negative impact on the average performance of Alpaca ($\overline{\Delta}_{\text{Alpaca}-both} = -0.1\%$, 2/6 tasks positive) or GPT-4 ($\overline{\Delta}_{\text{GPT-4}-both} = -0.3\%$, 3/6 tasks positive).
\review{This indicates that larger language models (LLaMA2, GPT-3.5 \vs Alpaca)} have a strong capability to leverage the information embedded in the prompts.
But for the huge GPT-4, adding prompts seems less effective, probably because it already contains similar basic information in its knowledge space.

We summarize our \textbf{key takeaways} from this section:
\textbf{
\begin{s_itemize}
\item Both small-scale and large-scale LLMs show promising performance on mental health tasks. FLAN-T5 and GPT-4's performance is approaching task-specific NLP models.
\item The prompt design enhancement strategies are generally effective for dialogue-focused models, but not for task-solving-focused models. These strategies work better for critical action prediction tasks such as suicide prediction.
\item Providing more contextual information about the task \& input can consistently improve performance in most cases.
\item Dialogue-focused models with larger trainable parameters (Alpaca \vs Alpaca-LoRA, as well as \review{LLaMA2/GPT-3.5 \vs Alpaca}) can better leverage the contextual or domain information in the prompts, yet GPT-4 shows less effect in response to different prompts.
\end{s_itemize}
}

\subsection{Few-shot Prompting Improves Performance to Some Extent}
\label{sub:results:few-shot}
We then investigate the effectiveness of few-shot prompting.
Note that since we observe diverse effects of prompt design strategies in Table~\ref{tab:results_zeroshot_delta}, in this section, we only experiment with the prompts with the best performance in the zero-shot setting. \review{Moreover, we exclude Alpaca-LoRA due to its less promising results and LLaMA2 due to its high computation cost.}

\begin{table}[!b]
\centering
\caption{
Balanced Accuracy Performance Change with Few-shot Prompting. This table is calculated between the zero-shot and the few-shot sections of Table~\ref{tab:results_overall_short}.
}
\label{tab:results_fewshot_delta}
\resizebox{0.64\textwidth}{!}{
\begin{tabular}{lcccc}
\thickhlinespace
\makecell[r]{\textbf{Dataset}} & \textbf{Dreaddit} & \multicolumn{2}{c}{\textbf{DepSeverity}} & \\ \addlinespace[1ex]
\textbf{Model}   & \textbf{Task \#1}       & \textbf{Task \#2}             & \textbf{Task \#3}     & $\overline{\Delta}-$All Three Tasks      \\ \thickhlinespace

$\Delta-$Alpaca$_{FS\_vs\_ZS}$ & \upgreen{ $\uparrow +0.039 $} & \upgreen{ $\uparrow +0.007 $} & \upgreen{ $\uparrow +0.197 $} & \upgreen{ $\uparrow +0.081 $}\\
$\Delta-$FLAN-T5$_{FS\_vs\_ZS}$ & \upgreen{$\uparrow$  +0.127} & \upgreen{$\uparrow +0.014 $} & \upgreen{ $\uparrow +0.036 $} & \upgreen{ $\uparrow +0.059 $}\\
$\Delta-$GPT-3.5$_{FS\_vs\_ZS}$ & \upgreen{ $\uparrow +0.036 $} & \upgreen{ $\uparrow +0.023 $} & \downred{ $\downarrow -0.023$} & \upgreen{ $\uparrow +0.012 $}\\
$\Delta-$GPT-4$_{FS\_vs\_ZS}$ & \downred{ $\downarrow -0.002$} & \upgreen{$\uparrow +0.005 $} & \upgreen{$\uparrow +0.025 $} & \upgreen{ $\uparrow +0.009 $}\\ \hdashlinespace
$\overline{\Delta}-$All Four Models & \upgreen{$\uparrow +0.051 $} & \upgreen{$\uparrow +0.012 $} & \upgreen{$\uparrow +0.059 $} & \upgreen{$\uparrow +0.041 $} \\

\thickhlinespace
\end{tabular}
}
\end{table}

Due to the maximum input token length of models (2048), we focus on Dreaddit and DepSeverity datasets that have a shorter input and experiment with $M=2$ in Eq.~\ref{eq:prompt-fs} for binary classification and $M=4/5$ for multi-class classification, \ie one sample per class. We repeat the experiment on each task three times and randomize the few shot samples for each run.

We summarize the overall results in the second section of Table~\ref{tab:results_overall_short} and the zoom-in comparison results in Table~\ref{tab:results_fewshot_delta}. 
Although language models with few-shot prompting still underperform task-specific models, providing examples of the task can improve model performance on most tasks compared to zero-shot prompting ($\overline{\Delta}_{FS\text{\_}vs\text{\_}ZS} = 4.1\%$).
Interestingly, few-shot prompting is more effective for Alpaca$_{FS}$ and FLAN-T5$_{FS}$ ($\overline{\Delta}_{\text{Alpaca}} = 8.1\%$, 3/3 tasks positive; $\overline{\Delta}_{\text{FLAN-T5}} = 5.9\%$, 3/3 tasks positive) than GPT-3.5$_{FS}$ and GPT-4$_{FS}$ ($\overline{\Delta}_{\text{GPT-3.5}} = 1.2\%$, 2/3 tasks positive; $\overline{\Delta}_{\text{GPT-4}} = 0.9\%$, 2/3 tasks positive).
Especially for Task \#3, we observe an improved balanced accuracy of 19.7\% for Alpaca but a decline of 2.3\% for GPT-3.5, so that Alpaca outperforms GPT-3.5 on this task. A similar situation is observed for FLAN-T5 (improved by 12.7\%) and GPT-4 (declined by 0.2\%) on Task \#1.
This may be attributed to the fact that smaller models such as Alpaca and FLAN-T5 can quickly adapt to complex tasks with only a few examples. In contrast, larger models like GPT-3.5 and GPT-4, with their extensive ``in memory'' data, find it more challenging to rapidly learn from new examples.

This leads to the key message from this experiment:
\textbf{Few-shot prompting can improve the performance of LLMs on mental health prediction tasks to some extent, especially for small models.} 

\subsection{Instruction Finetuning Boost Performance for Multiple Tasks Simultaneously}
\label{sub:results:finetune}
Our experiments so far have shown that zero-shot and few-shot prompting can improve LLMs on mental health tasks to some extent, but their overall performance is still below state-of-the-art task-specific models.
In this section, we explore the effectiveness of instruction finetuning.

\begin{table}[!b]
\centering
\caption{
Balanced Accuracy Performance Change with Instruction Finetuning. This table is calculated between the finetuning and zero-shot section, as well as the finetuning and few-shot section of Table~\ref{tab:results_overall_short}. GPT-3.5$_{Best}$ and GPT-4$_{Best}$ are the best results among zero-shot and few-shot settings.
}
\label{tab:results_finetuning_delta}
\resizebox{0.73\textwidth}{!}{
\begin{tabular}{lcccccc}
\thickhlinespace
\makecell[r]{\textbf{Dataset}} & \textbf{Dreaddit} & \multicolumn{2}{c}{\textbf{DepSeverity}} & \textbf{SDCNL} & \multicolumn{2}{c}{\textbf{CSSRS-Suicide}} \\ \addlinespace[1ex]
\textbf{Model}   & \textbf{Task \#1}       & \textbf{Task \#2}             & \textbf{Task \#3}             & \textbf{Task \#4}    & \textbf{Task \#5}     & \textbf{Task \#6} \\ \thickhlinespace
$\Delta-$Alpaca$_{FT\_vs\_ZS}$ & \upgreen{ $\uparrow +0.223$} & \upgreen{ $\uparrow +0.253 $} & \upgreen{ $\uparrow +0.315 $} & \upgreen{ $\uparrow +0.231 $} & \upgreen{ $\uparrow +0.212 $} & \upgreen{ $\uparrow +0.171 $} \\
$\Delta-$Alpaca$_{FT\_vs\_FS}$ & \upgreen{ $\uparrow +0.184 $} & \upgreen{ $\uparrow +0.246 $} & \upgreen{ $\uparrow +0.118 $} & --- & --- & --- \\
\hdashlinespace
$\Delta-$FLAN-T5$_{FT\_vs\_ZS}$ & \upgreen{ $\uparrow +0.143 $} & \upgreen{ $\uparrow +0.095 $} & \upgreen{ $\uparrow +0.360 $} & \upgreen{ $\uparrow +0.047 $} & \upgreen{ $\uparrow +0.201 $} & \upgreen{ $\uparrow +0.034 $}  \\
$\Delta-$FLAN-T5$_{FT\_vs\_FS}$ & \upgreen{ $\uparrow +0.016 $} & \upgreen{ $\uparrow +0.081 $} & \upgreen{ $\uparrow +0.324 $} & --- & --- & --- \\
\thickhlinespace
\textit{Results comparison:} & &&&&&\\
Mental-RoBERTa   & \textbf{0.831}  & \textbf{0.790}  & 0.736  & \underline{0.723}  & \underline{0.853}  & 0.373\\
GPT-3.5$_{Best}$ & 0.721 & 0.665 & 0.642 & 0.632 & 0.617 & 0.310\\
GPT-4$_{Best}$ & 0.725 & 0.724 & 0.656 & 0.647 & 0.760 & \underline{0.441}\\
Mental-Alpaca    & \underline{0.816}  & \underline{0.775} & \underline{0.746} & \textbf{0.724} & 0.730 & 0.403 \\
Mental-FLAN-T5 &  0.802 & 0.759 & \textbf{0.756} & 0.677 & \textbf{0.868} & \textbf{0.481} \\
\thickhlinespace
\end{tabular}
}
\end{table}

Due to the prohibitive cost and lack of transparency of GPT-3.5 and GPT-4 finetuning, we only experiment with Alpaca and FLAN-T5 that we have full control of.
We picked the most informative prompt to provide more embedded knowledge during the finetuning.
As introduced in Sec.~\ref{subsub:methods:finetuning:multi-dataset} and Sec.~\ref{subsub:implementation:models:mental}, we build Mental-Alpaca and Mental-FLAN-T5 by finetuning Alpaca and FLAN-T5 on all six tasks across four datasets at the same time.

The third section of Table~\ref{tab:results_overall_short} summarizes the overall results, and Table~\ref{tab:results_finetuning_delta} highlights the key comparisons.
We observe that both Mental-Alpaca and Mental-FLAN-T5 achieve significantly better performance compared to the unfinetuned versions ($\overline{\Delta}_{\text{Alpaca}-FT\_vs\_ZS}$ = 23.4\%, $\overline{\Delta}_{\text{Alpaca}-FT\_vs\_FS}$ = 18.3\%; $\overline{\Delta}_{\text{FLAN-T5}-FT\_vs\_ZS}$ = 14.7\%, $\overline{\Delta}_{\text{FLAN-T5}-FT\_vs\_FS}$ = 14.0\%).
Both finetuned models surpass GPT-3.5's best performance among zero-shot and few-shot settings across all six tasks ($\overline{\Delta}_{\text{Mental-Alpaca}\_vs\_\text{GPT-3.5}}$ = 10.1\%; $\overline{\Delta}_{\text{Mental-FLAN-T5}\_vs\_\text{GPT-3.5}}$ = 11.6\%) and outperform GPT-4's best version in most tasks  ($\overline{\Delta}_{\text{Mental-Alpaca}\_vs\_\text{GPT-4}}$ = 4.0\%, 4/6 tasks positive; $\overline{\Delta}_{\text{Mental-FLAN-T5}\_vs\_\text{GPT-4}}$ = 5.5\%, 5/6 tasks positive).
Recall that GPT-3.5/GPT-4 are 25/250 times bigger than Mental-Alpaca and 15/150 times bigger than Mental-FLAN-T5.

More importantly, Mental-Alpaca and Mental-FLAN-T5 perform on par with the state-of-the-art Mental-RoBERTa. Mental-Alpaca has the best performance on one task and the second best on three tasks, while Mental-FLAN-T5 has the best performance on three tasks. It is noteworthy that Mental-RoBERTa is a task-specific model, which means it is specialized on one task after being trained on it. In contrast, Mental-Alpaca and Mental-FLAN-T5 can simultaneously work across \textit{all} tasks with a single-round finetuning.
These results show the strong effectiveness of instruction finetuning: By finetuning LLMs on multiple mental health datasets with instructions, the models can obtain better capability to solve a variety of mental health prediction tasks.

\subsubsection{Dialogue-Focused \vs Task-Solving-Focused LLMs}
\label{subsub:results:finetune:dialogue_vs_tasksolving}
We further compare Mental-Alpaca and Mental-FLAN-T5.
Overall, their performance is quite close ($\overline{\Delta}_{\text{FLAN-T5}\_vs\_\text{Alpaca}} = 1.4\%$), with Mental-Alapca better at Task \#4 on SDCNL and Mental-FLAN-T5 better at Task \#5 and \#6 on CSSRS-Suicide.
In Sec.~\ref{sub:results:zero-shot}, we observe that FLAN-T5$_{ZS}$ has a much better performance than Alpaca$_{ZS}$ ($\overline{\Delta}_{\text{FLAN-T5}\_vs\_\text{Alpaca}}$ = 10.9\%, 5/6 tasks positive) in the zero-shot setting. However, after finetuning, FLAN-T5's advantage disappears.

Our comparison result indicates that Alpaca, as a dialogue-focused model, is better at learning from human natural language data compared to FLAN-T5. Although FLAN-T5 is good at task solving and thus has a better performance in the zero-shot setting, its performance improvement after instruction finetuning is relatively smaller than that of Alpaca.
This observation has implications for future stakeholders. If the data and computing resources for finetuning are not available, using task-solving-focused LLMs could lead to better results. When there are enough data and computing resources, finetuning dialogue-based models can be a better choice.
Furthermore, models like Alpaca, with their dialogue conversation capabilities, may be more suitable for downstream applications, such as mental well-being assistants for end-users.

\subsubsection{Does Finetuning Generalize across Datasets?}
\label{subsub:results:finetune:generalizability}
We further measure the generalizability of LLMs after finetuning. To do this, we instruction-finetune the model on one dataset and evaluate it on all datasets (as introduced in Sec.~\ref{subsub:methods:finetuning:single-dataset}). 
As the main purpose of this part is not to compare different models but evaluate the finetuning method, we only focus on Alpaca.
Table~\ref{tab:results_transfer} summarizes the results.

\begin{table}[b]
\centering
\caption{
Balanced Accuracy Cross-Dataset Performance Summary of Mental-Alpaca Finetuning on Single Dataset.
\fbox{Numbers} indicate the results of the model finetuned and tested on the same dataset.
The bottom few rows are related Alpaca versions for reference. \upgreen{$\uparrow$}/\downred{$\downarrow$} marks the ones with better/worse cross-dataset performance compared to the zero-shot version Alpaca$_{ZS}$.
}
\vspace{-0.3cm}
\label{tab:results_transfer}
\resizebox{0.75\textwidth}{!}{
\begin{tabular}{lcccccc}
\thickhlinespace
\makecell[r]{\textbf{Test Dataset}} & \textbf{Dreaddit} & \multicolumn{2}{c}{\textbf{DepSeverity}} & \textbf{SDCNL} & \multicolumn{2}{c}{\textbf{CSSRS-Suicide}} \\ \addlinespace[1ex]
\textbf{Finetune Dataset}   & \textbf{Task \#1}       & \textbf{Task \#2}             & \textbf{Task \#3}             & \textbf{Task \#4}    & \textbf{Task \#5}     & \textbf{Task \#6} \\ \thickhlinespace
Dreaddit & \fbox{0.823} & \upgreen{ $\uparrow$}  0.720 & \upgreen{ $\uparrow$} 0.623                      & \downred{ $\downarrow$} 0.474 & \upgreen{ $\uparrow$} 0.720                      & \downred{ $\downarrow$}  0.156                      \\ 
DepSeverity                   & \upgreen{ $\uparrow$}  0.618 & \fbox{0.733} & \fbox{0.769} & $|$ 0.493 & \upgreen{ $\uparrow$} 0.753                      & \downred{ $\downarrow$} 0.156                      \\ 
SDCNL                         & \downred{ $\downarrow$} 0.468                      & \downred{ $\downarrow$} 0.461 & \upgreen{ $\uparrow$} 0.623 & \fbox{0.730} & \upgreen{ $\uparrow$} 0.573 & \downred{ $\downarrow$} 0.156                      \\
CSSRS-Suicide                 & \downred{ $\downarrow$} 0.500                      & \downred{ $\downarrow$} 0.500 & \upgreen{ $\uparrow$} 0.622                      & \upgreen{ $\uparrow$}  0.500 & \fbox{0.753} & \fbox{0.578} \\  \hdashlinespace
\textit{Reference:} & &&&&&\\
Alpaca$_{ZS}$       & 0.593 & 0.522 & 0.431 & 0.493 & 0.518 & 0.232 \\ 
Mental-Alpaca    & 0.816 & 0.775 & 0.746 & 0.724 & 0.730 & 0.403 \\
\thickhlinespace
\end{tabular}
}
\end{table}

We first find that finetuning and testing on the same dataset lead to good performance, as indicated by the \fbox{boxed} entries on the diagonal in Table~\ref{tab:results_transfer}. Some results are even better than Mental-Alpaca (5 out of 6 tasks) or Mental-RoBERTa (3 out of 6 tasks), which is not surprising.
More interestingly, we investigate cross-dataset generalization performance (\ie the ones off the diagonal).
Overall, finetuning on a single dataset achieves better performance compared to the zero-shot setting ($\overline{\Delta}_{FT\text{-Single}\_vs\_ZS} = 4.2\%$).
However, the performance changes vary across tasks. For example, finetuning on any dataset is beneficial for Task \#3 ($\overline{\Delta} = 19.2\%$) and \#5 ($\overline{\Delta} = 16.4\%$), but detrimental for Task \#6 ($\overline{\Delta} = -7.6\%$) and almost futile for Task \#4 ($\overline{\Delta} = -0.4\%$). Generalizing across Dreaddit and DepSeverity shows good performance, but this is mainly because they share the language corpus.
These results indicate that finetuning on a single dataset can provide mental health knowledge with a certain level and thus improve the overall generalization results, but such improvement is not stable across tasks.

\begin{table}[!t]
\centering
\caption{
\review{
Balanced Accuracy Performance Summary on Three External Datasets. These datasets come from diverse social media platforms.
For each column, the best result is \textbf{bolded}, and the second best is \underline{underlined}.
}}
\label{tab:results_external_overall}
\resizebox{0.6\textwidth}{!}{
\begin{tabular}{llccc}
\thickhlinespace
& \makecell[r]{\textbf{Dataset}} & \textbf{Red-Sam} & \textbf{Twt-60Users} & \textbf{SAD} \\ \addlinespace[1ex]
\textbf{Category} & \textbf{Model}   & \textbf{Task \#2}       & \textbf{Task \#2}             & \textbf{Task \#1}\\ \thickhlinespace
\multirow{6}{*}{\makecell{Zero-shot\\Prompting}}  
 & Alpaca$_{ZS\_best}$ & 0.527$_{\pm0.006}$ & 0.569$_{\pm0.017}$ & 0.557$_{\pm0.041}$\\ 
 & Alpaca-LoRA$_{ZS\_best}$ & 0.577$_{\pm0.004}$ & 0.649$_{\pm0.021}$ & 0.477$_{\pm0.016}$\\ 
 & FLAN-T5$_{ZS\_best}$ & 0.563$_{\pm0.029}$ & 0.613$_{\pm0.046}$ & 0.767$_{\pm0.050}$\\ 
 & LLaMA2$_{ZS\_best}$ & 0.574$_{\pm0.008}$ & \underline{0.736}$_{\pm0.019}$ & 0.704$_{\pm0.026}$\\ 
 & GPT-3.5$_{ZS\_best}$ & 0.506$_{\pm0.004}$ & 0.571$_{\pm0.000}$ & 0.750$_{\pm0.027}$\\ 
 & GPT-4$_{ZS\_best}$ & 0.511$_{\pm0.000}$ & 0.566$_{\pm0.017}$ & \textbf{0.854}$_{\pm0.006}$\\
\thickhlinespace
\multirow{5}{*}{\makecell{Instructional\\Finetuning}} & Mental-Alpaca         & \textbf{0.604}$_{\pm0.012}$ & 0.718$_{\pm0.011}$ & \underline{0.819}$_{\pm0.006}$ \\
 & $\Delta-$Alpaca$_{FT\_vs\_ZS}$ & \upgreen{ $\uparrow +0.077$} & \upgreen{ $\uparrow +0.149 $} & \upgreen{ $\uparrow +0.262 $} \\\\
& Mental-FLAN-T5 & \underline{0.582}$_{\pm0.002}$ & \textbf{0.736}$_{\pm0.003}$ & 0.779$_{\pm0.002}$ \\
 & $\Delta-$FLAN-T5$_{FT\_vs\_ZS}$ & \upgreen{ $\uparrow +0.019$} & \upgreen{ $\uparrow +0.123 $} & \upgreen{ $\uparrow +0.012 $}\\
\thickhlinespace
\end{tabular}
}
\end{table}

\review{
Moreover, we further evaluate the generalizability of our best models instructional-finetuned on multiple datasets, \ie Mental-Alpaca and Mental-FLAN-T5. We leverage external datasets that are not included in the finetuning. Table~\ref{tab:results_external_overall} highlights the key results. More detailed results can be found in Table~\ref{tab:results_overall}.
}

\review{
Consistent with the results in Table~\ref{tab:results_finetuning_delta}, the instruction finetuning enhances the model performance on external datasets ($\overline{\Delta}_{Alpaca} = 16.3\%$, $\overline{\Delta}_{FLAN-T5} = 5.1\%$). Both Mental-Alpaca and Mental-FLAN-T5 ranked top 1 or 2 in 2/3 external tasks.
It is noteworthy that Twt-60Users and SAD datasets are collected outside Reddit, and their data is different from the source of finetuning datasets. These results demonstrate strong evidence that instruction finetuning with diverse tasks, even with data collected from a single social media platform, can significantly enhance LLMs' generalizability across multiple scenarios.
}

\subsubsection{How Much Data Is Needed?}
\label{subsub:results:finetune:datasetsize}
\begin{figure}[t]
    \centering
    \includegraphics[width=1\columnwidth]{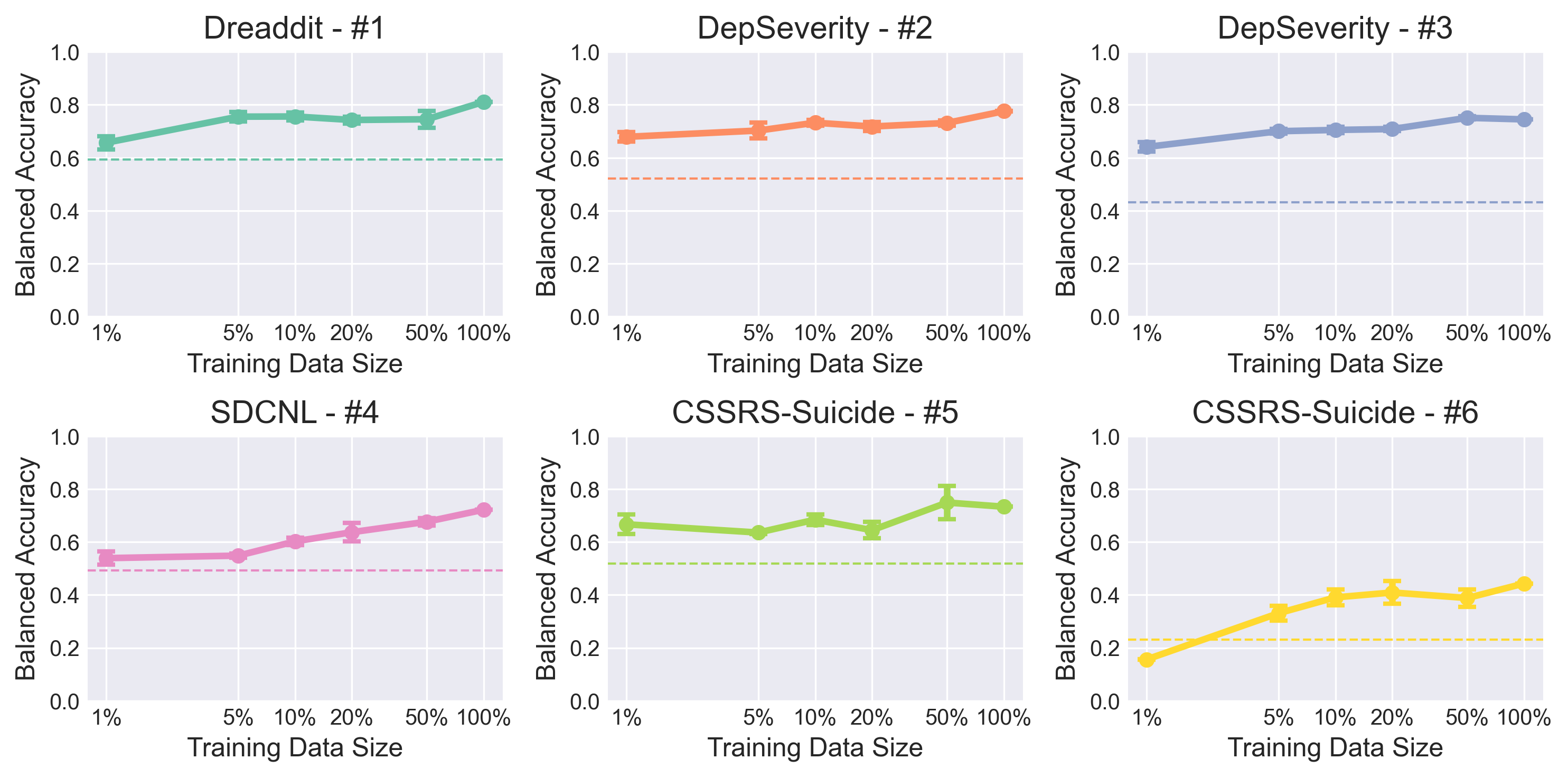}
    \caption{Balanced Accuracy Performance Summary of Mental-Alpaca Finetuning with Different Sizes of Training Set.
    The finetuning is conducted across four datasets and six tasks. Each solid line represents the performance of the finetuned model on each task. The dashed line indicates the Alpaca$_{ZS}$ performance baseline.
    Note that the x-axis is in the log scale.
    }
    \label{fig:performance_datasize}
\end{figure}

Additionally, we are interested in exploring how the size of the dataset impacts the results of instruction finetuning. To answer this question, we downsample the training set to 50\%, 20\%, 10\%, 5\%, and 1\% of the original size and repeat each one three times. We increase the training epoch accordingly to make sure that the model is exposed to a similar amount of data.
Similarly, we also focus on Alpaca only.
Figure~\ref{fig:performance_datasize} visualizes the results.
With only 1\% of the data, the finetuned model is able to outperform the zero-shot model on most tasks (5 out of 6). With 5\% of the data, the finetuned model has a better performance on all tasks.
As expected, the model performance has an increasing trend with more training data. For many tasks, the trend approaches a plateau after 10\%. The difference between 10\% training data (less than 300 samples per dataset) and 100\% training data is not huge ($\overline{\Delta}$ = 5.9\%).

\subsubsection{More Data in One Dataset \vs Fewer Data across Multiple Datasets}
In Sec.~\ref{subsub:results:finetune:generalizability}, the finetuning on a single dataset can be viewed as training on a smaller set (around 5-25\% of the original size) with less variation (\ie no finetuning across datasets). Thus, the results in Sec.~\ref{subsub:results:finetune:generalizability} are comparable to those in Sec.~\ref{subsub:results:finetune:datasetsize}.
We found that the model's overall performance is better when the model is finetuned across multiple datasets when overall training data sizes are similar ($\overline{\Delta}_{FT\text{-}5\%\_vs\_FT\text{-Single}}$ = 3.8\%, $\overline{\Delta}_{FT\text{-}10\%\_vs\_FT\text{-Single}}$ = 8.1\%, $\overline{\Delta}_{FT\text{-}20\%\_vs\_FT\text{-Single}}$ = 12.4\%).
This suggests that increasing data variation can more effectively benefit finetuning outcomes when the training data size is fixed.

These results can guide future developers and practitioners in collecting the appropriate data size and sources to finetune LLMs for the mental health domain efficiently. We have more discussion in the next section.
In summary, we highlight the \textbf{key takeaways} of our finetuning experiments as follows:
\textbf{
\begin{s_itemize}
\item Instruction finetuning on multiple mental health datasets can significantly boost the performance of LLMs on various mental health prediction tasks. Mental-Alpaca and Mental-FLAN-T5 outperform GPT-3.5 and GPT-4, and perform on par with the state-of-the-art task-specific model.
\item Although task-solving-focused LLMs may have better performance in the zero-shot setting for mental health prediction tasks, dialogue-focused LLMs have a stronger capability of learning from human natural language and can improve more significantly after finetuning.
\item \review{Finetuning LLMs on a small number of datasets and tasks may improve model generalizable knowledge in mental health, but its effect is not robust. Comparatively, finetuning on diverse tasks can robustly enhance generalizability across multiple social media platforms.}
\item Finetuning LLMs on a small number of samples (a few hundred) across multiple datasets can already achieve favorable performance.
\item When the data size is the same, finetuning LLMs on data with larger variation (\ie more datasets and tasks) can achieve better performance.
\end{s_itemize}
}

\subsection{Case Study of LLMs' Capability on Mental Health Reasoning}
\label{sub:results:reasoning}

\begin{figure}[b]
    \centering
    \includegraphics[width=1\columnwidth]{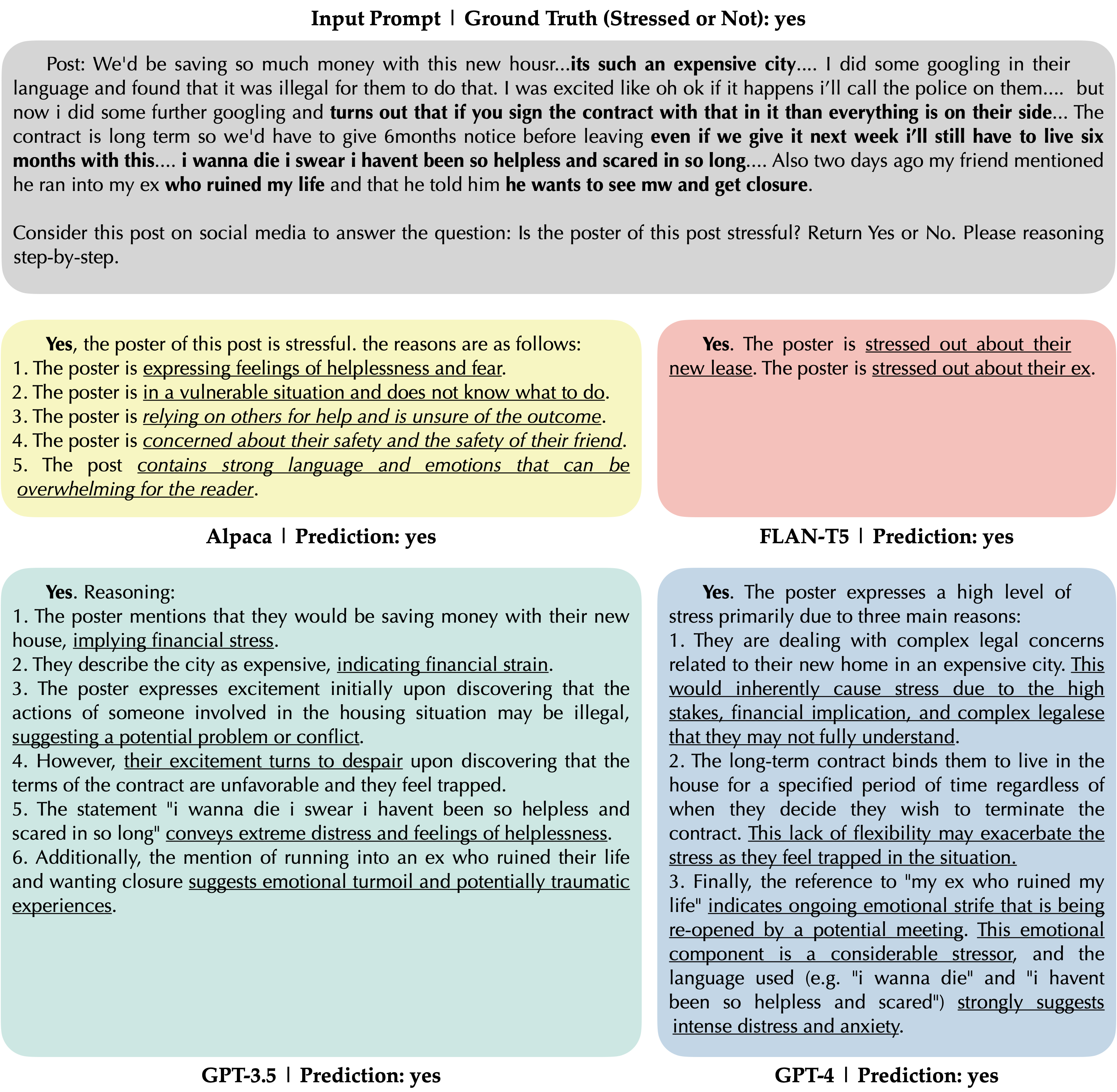}
    \caption{A Case Study of Correct Reasoning Examples on Task \#1 Binary Stress Prediction on Dreaddit Dataset. \textbf{Bolded texts} highlight the mental-health-related content in the input section, and the answers of LLMs. \underline{Underlined texts} highlight the reasoning content generated by LLMs, and \textit{\underline{italicized \& underlined texts}} indicate the \textit{wrong} or unrelated content.
    }
    \label{fig:reasoning_example1}
\end{figure}

In addition to evaluating LLMs' performance on classification tasks, we also take an initial step to explore LLMs' capability on mental health reasoning.
This is another strong advantage of LLMs since they can generate human-like natural language based on embedded knowledge.
Due to the high cost of a systematic evaluation of reasoning outcomes, here we present a few examples as a case study across different LLMs.
It is noteworthy that we do not aim to claim that certain LLMs have better/worse reasoning capabilities. Instead, this section aims to provide a general sense of LLMs' performance on mental health reasoning tasks.

Specifically, we modify the prompt design by inserting a Chain-of-Thought (CoT) prompt~\cite{kojima_large_2022} at the end of \textit{OutputConstraint} in Eq.~\ref{eq:prompt-zs}: ``Return [set of classification labels]. Provide reasons step by step''. We compare Alpaca, FLAN-T5, GPT-3.5, and GPT-4. Our results indicate the promising reasoning capability of these models, especially GPT-3.5 and GPRT-4.
We also experimented with the finetuned Mental-Alpaca and Mental-FLAN-T5. Unfortunately, our results show that after finetuning solely on classification tasks, these two models are no longer able to generate reasoning sentences even with the CoT prompt. This suggests a limitation of the current finetuned model.

\subsubsection{Diverse Reasoning Capabilities across LLMs.}
\label{subsub:results:reasoning:okay}
\review{
We present several examples as our case study to illustrate the reasoning capability of these LLMs.
}
The first example comes from the binary stress prediction task (Task \#1) on the Dreaddit dataset (see Figure~\ref{fig:reasoning_example1}).
All models give the right classification, but with significantly different reasoning capabilities.
First, FLAN-T5 generates the shortest reason. Although it is reasonable, it is superficial and does not provide enough insights. This is understandable because FLAN-T5 is targeted at task-solving instead of reasoning.
Compared to FLAN-T5, Alpaca generates better reason. Among the five reasons, two of them accurately analyze the user's mental state given the stressful situations.
Meanwhile, GPT-3.5 and GPT-4 generate expert-level high-quality reasons. The inference from the user's statement is accurate and deep, indicating their powerful capability of understanding human emotion and mental health. Comparing the two models, GPT-3.5's reason is simpler, following the user's statement point by point and adding basic comments, while GPT-4's output is more organic and insightful, yet more concise.

\begin{figure}[t]
    \centering
    \includegraphics[width=1\columnwidth]{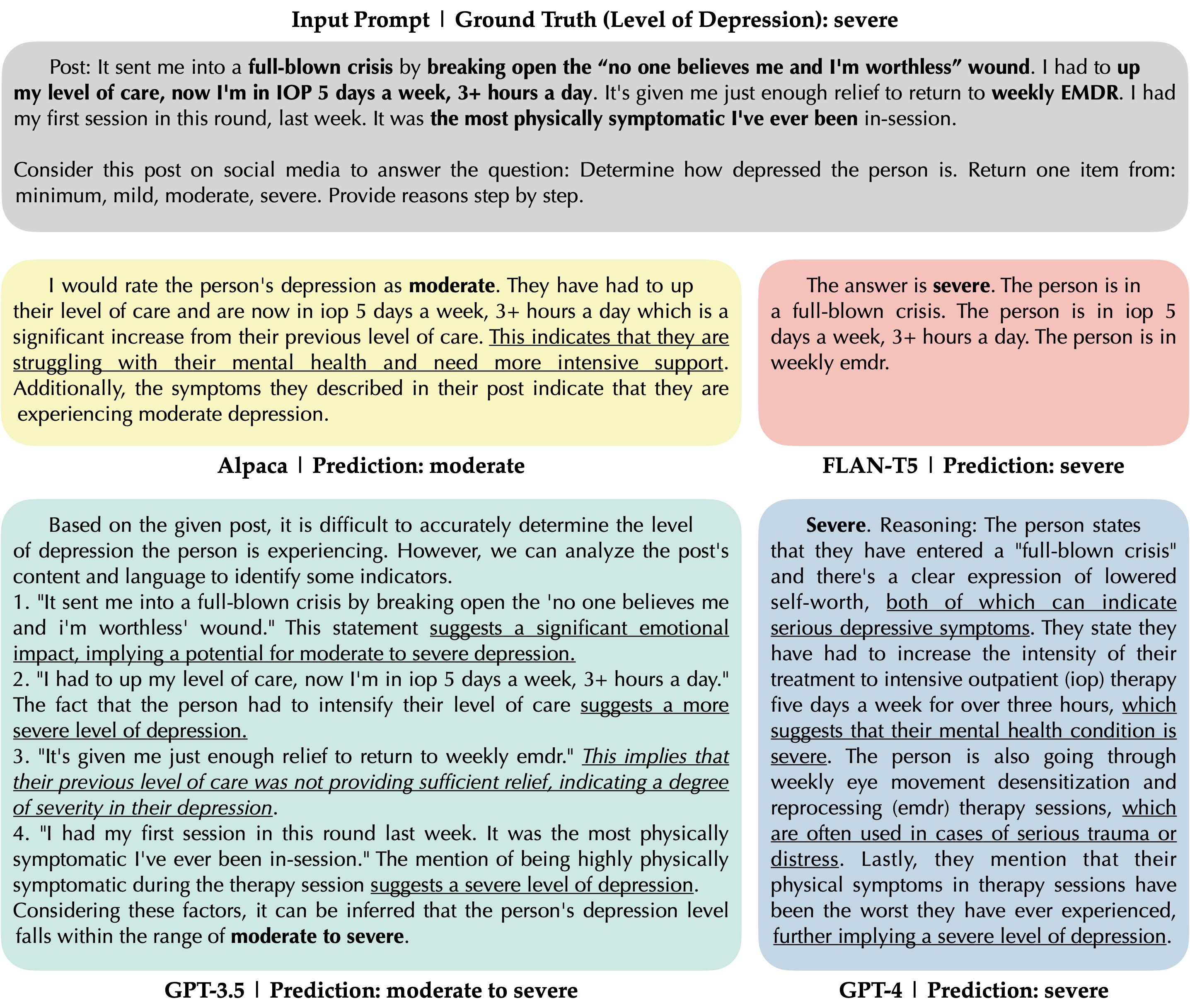}
    \caption{A Case Study of Mixed Reasoning Examples on Task \#3  Four-level Depression Prediction on DepSeverity Dataset. Alpaca wrongly predicted the label, and GPT-3.5 provided a wrong inference on the meaning of ``relief''.
    }
    \label{fig:reasoning_example2}
\end{figure}

We also have a similar observation in the second example from the four-level depression prediction task (Task \#3) on the DepSeverity dataset (see Figure~\ref{fig:reasoning_example2}).
In this example, although FLAN-T5's prediction is correct, it simply repeats the fact stated by the user, without providing further insights.
Alpaca makes the wrong prediction, but it provides one sentence of accurate reasoning (although relatively shallow).
GPT-3.5 makes an ambiguous prediction that includes the correct answer.
In contrast, GPT-4 generates the highest quality reasoning with the right prediction. With its correct understanding of depressive symptoms, GPT-4 can accurately infer from the user's situation, link it to symptoms, and provides insightful analysis.

\subsubsection{Wrong and Dangerous Reasoning from LLMs.}
\label{subsub:results:reasoning:wrong}
\begin{figure}[t]
    \centering
    \vspace{-0.5cm}
    \includegraphics[width=1\columnwidth]{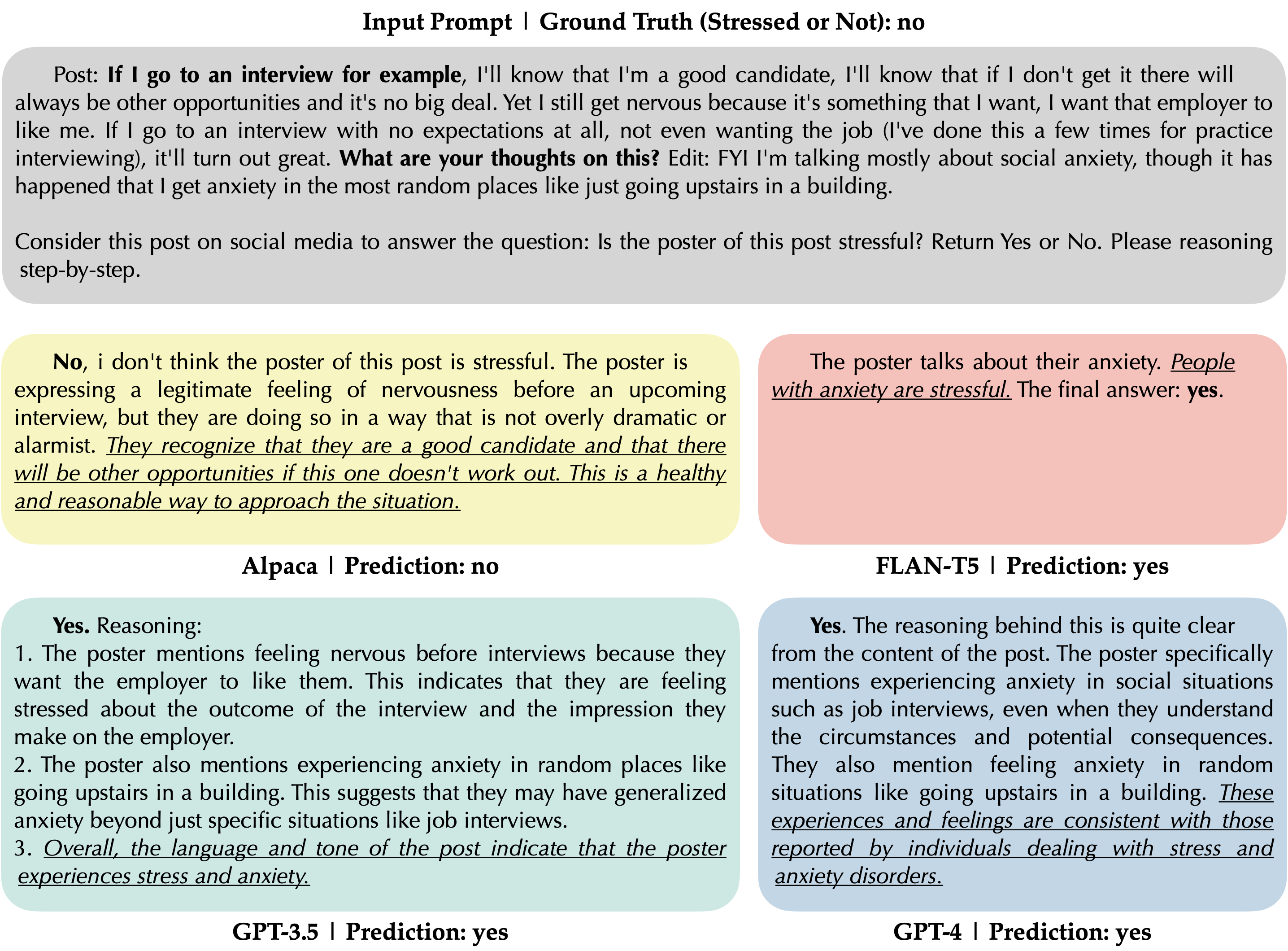}
    \vspace{-0.5cm}
    \caption{A Case Study of \textbf{\textit{Incorrect}} Reasoning Examples on Task \#1 Binary Stress Prediction on Dreaddit Dataset. FLAN-T5, GPT-3.5, and GPT-4 all make false positive predictions. All four LLMs provide problematic reasons.
    }
    \label{fig:reasoning_example3}
    \vspace{-0.5cm}
\end{figure}

\review{
However, we also want to emphasize the \textit{incorrect} reasoning content, which may lead to negative consequences and risks.
}
In the first example, Alpaca generated two wrong reasons for the hallucinated ``reliance on others'' and ``safety concerns'', along with an unrelated reason for readers instead of the poster. In the second example, GPT-3.5 misunderstood what the user meant by ``relief''. To better illustrate this, we further present another example where all four LLMs make problematic reasoning (see Figure~\ref{fig:reasoning_example3}).
In this example, the user was asking for others' opinions on social anxiety, with their own job interview experience as an example. Although the user mentioned situations where they were anxious and stressed, it's clear that they were calm when writing this post and described their experience in an objective way. However, FLAN-T5, GPT-3.5, and GPT-4 all mistakenly take the description of the anxious interview experience as evidence to support their wrong prediction. Although Alpaca makes the right prediction, it does not understand the main theme of the post. The false positives reveal that LLMs may overly generalize in a wrong way: Being stressed in one situation does not indicate that a person is stressed all the time.
However, the reasoning content alone reads smoothly and logically. If the original post was not provided, the content could be very misleading, resulting in a wrong prediction with reasons that ``appears to be solid''.
These examples clearly illustrate the limitations of the current LLMs for mental health reasons, as well as their risks of introducing dangerous bias and negative consequences to users.

The case study suggests that GPT-4 enjoys impressive reasoning capability, followed by GPT-3.5 and Alpaca. Although FLAN-T5 shows a promising zero-shot performance, it is not good at reasoning.
Our results reveal the encouraging capability of LLMs to understand human mental health and generate meaningful analysis.
However, we also present examples where LLMs can make mistakes and offer explanations that appear reasonable but are actually flawed. This further suggests the importance of more future research on LLMs' ethical concerns and safety issues before real-world deployment.

\section{Discussion}
\label{sec:discussion}
Our experiment results reveal a number of interesting findings. In this section, we discuss potential guidelines for enabling LLMs for mental -health-related tasks (Sec.~\ref{sub:discussion:guidelines}). We envision promising future directions (Sec.~\ref{sub:discussion:beyond}), while highlighting important ethical concerns and limitations with LLMs for mental health (Sec.~\ref{sub:discussion:ethics}). We also summarize the limitations of the current work (Sec.~\ref{sub:discussion:limitation}).

\subsection{Guidelines for Empowering LLMs for Mental Health Prediction Tasks}
\label{sub:discussion:guidelines}
We extract and summarize the takeaways from Sec.~\ref{sec:results} into a set of guidelines for future researchers and practitioners on how to empower LLMs to be better at various mental health prediction tasks.

\textbf{When computing resources are limited, combine prompt design \& few-shot prompting, and pick prompts carefully.}
As the size of large models continues to grow, the requirement for hardware (mainly GPU) has also been increasing, especially when finetuning an LLM. For example, in our experiment, Alpaca was trained on eight 80GB A100s for three hours~\cite{taori_stanford_2023}.
With limited computing resources, only running inferences or resorting to APIs is feasible. In these cases, zero-shot and few-shot prompt engineering strategies are viable options. Our results indicate that providing few-shot mental health examples with appropriate enhancement strategies can effectively improve prediction performance.
Specifically, adding contextual information about the online text data is always helpful. If the available model is large and contains rich knowledge (at least 7B trainable parameters), adding mental health domain information can also be beneficial.


\textbf{With enough computing resources, instruction finetune models on various mental health datasets.}
When there are enough computing resources and model training/finetuning is possible, there are more options to enhance LLMs for mental health prediction tasks.
Our experiments clearly show that instruction finetuning can significantly boost the performance of models, especially for dialogue-based models since they can better understand and learn from human natural language.
\review{When there are multiple datasets available, merging multiple datasets and tasks altogether and finetuning the model in a single round is the most effective approach to enhance its generalizability.
}

\textbf{Implement efficient finetuning with hundreds of examples and prioritize data variation when data resource is limited.}
Figure~\ref{fig:performance_datasize} shows that finetuning does not require large datasets. If there is no immediately available dataset, collecting small datasets with a few hundred samples is often good enough.
\review{
Moreover, when the overall amount of data is limited (\eg due to resource constraints), it is more advantageous to collect data from a variety of sources, each with a smaller size, than to collect a single larger dataset. Because instruction finetuning generalizes better when data and tasks have a larger variation. 
}


\textbf{More curated finetuning datasets are needed for mental health reasoning.}
Our case study suggests that Mental-Alpaca and Mental-FLAN-T5 can only generate classification labels after being finetuned solely on classification tasks, losing their reasoning capability. This is a major limitation of the current models.
A potential solution involves incorporating more datasets focused on reasoning or causality into the instruction finetuning process, so that models can also learn the relationship between mental health outcomes and causal factors.

\review{
\textbf{Limited Prediction and Reasoning Performance for Complex Contexts.}
LLMs tend to make more mistakes when the conversation contexts are more complex~\cite{bhattacharya2023context,li2023compressing}.
Our results contextualize this in the mental health domain.
Section~\ref{subsub:results:reasoning:wrong} shows an example case where most LLMs not only predict incorrectly but also provide flawed reasoning processes.
Further analysis of mispredicted instances indicates a recurring difficulty: LLMs often err when there's a disconnect between the literal context and the underlying real-life scenarios. 
he example in Figure~\ref{fig:reasoning_example3} is a case where LLMs are confused by the hypothetical stressful case described by the person.
In another example, all LLMs incorrectly assess a person with severe depression (false negative): ``\textit{I'm just blown away by this doctor's willingness to help. I feel so validated every time I leave his office, like someone actually understands what I'm struggling with, and I don't have to convince them of my mental illness. Bottom line? Research docs if you can online, read their reviews and don't give up until you find someone who treats you the way you deserve. If I can do this, I promise you can!}'' 
Here, LLMs are misled by the outwardly positive sentiment, overlooking the significant cues such as regular doctor visits and explicit mentions of mental illness.
These observations underscore a critical shortfall of LLMs: they cannot handle complex mental health-related tasks, particularly those concerning chronic conditions like depression. The variability of human expressions over time and the models' susceptibility to being swayed by superficial text rather than underlying scenarios present significant challenges.
}

Despite the promising capability of LLMs for mental health tasks, they are still far from being deployable in the real world.
Our experiments reveal the encouraging performance of LLMs on mental health prediction and reasoning tasks.
However, as we note in Sec.~\ref{sub:discussion:ethics}, our current results do not indicate LLMs' deployability in real-life mental health settings. There are many important ethical and safety concerns and gaps before deployment to be addressed before achieving robustness and deployability.

\subsection{Beyond Mental Health Prediction Task and Online Text Data}
\label{sub:discussion:beyond}
Our current experiments mainly involve mental health prediction tasks, which are essentially classification problems. There are more types of tasks that our experiments don't cover, such as regression (\eg predicting a score on a mental health scale). In particular, reasoning is an attractive task as it can fully leverage the capability of LLMs on language generation~\cite{nori_capabilities_2023,bubeck_sparks_2023}. Our initial case study on reasoning is limited but reveals promising performance, especially for large models such as GPT-4. We plan to conduct more experiments on tasks that go beyond classification.

In addition, there is another potential extension direction. In this paper, we mainly focus on online text data, which is one of the important data sources of the ubiquitous computing ecosystem.
\minorreview{However, there are more available data streams that contain rich information, such as the multimodal sensor data from mobile phones and wearables (\eg \cite{xu_globem_2022, nepal2022survey,xu_understanding_2021,morais2023classification,JAbrantes_2023}).}
This leads to another open question on how to leverage LLMs for time-series sensor data. More research is needed to explore potential methods to merge the online text information with sensor streams. These are another set of exciting research questions to explore in future work.

\subsection{Ethics in LLMs and Deployability Gaps for Mental Health}
\label{sub:discussion:ethics}
Although our experiments on LLMs have shown promising capability for mental-health-related tasks, it still has a long way to go before being deployed in real-life systems. Recent research has revealed the potential bias or even harmful advice introduced by LLMs~\cite{Hoover_2023}, especially with the gender~\cite{ghosh2023chatgpt} and racial~\cite{abid2021persistent} gaps.
In mental health, these gaps and disparities between population groups have been long-standing~\cite{irene_y_chen_can_2019}.
Meanwhile, our case study of incorrect prediction, over-generalization, and ``falsely reasonable'' explanations further highlight the risk of current LLMs.
Recent studies are calling for more research emphasis and efforts in assessing and mitigating these biases for mental health~\cite{irene_y_chen_can_2019,timmons2022call}.

Although with a much stronger capability of understanding natural language (and early signs of mental health domain knowledge in our case), LLMs are no different from other modern AI models that are trained on a large amount of human-generated content, which exhibit all the biases that humans do~\cite{irene_y_chen_ethical_2021,ntoutsi2020bias,wang2020human}.
Meanwhile, although we carefully picked datasets with human expert annotations, there exist potential biases in the labels, such as stereotypes~\cite{pessach2022review}, confirmation bias~\cite{gemalmaz2021accounting}, normative \vs descriptive labels~\cite{balagopalan_judging_2023}. 
Besides, privacy is another important concern. Although our datasets are based on public social media platforms, it is necessary to carefully handle mental-health-related data and guarantee anonymity in any future efforts.
These ethical concerns need to receive attention not only at the monitoring and prediction stage, but also in the downstream applications, ranging from assistants for mental health experts to chatbots for end-users.
Careful efforts into safe development, auditing, and regulation are very much needed to address these ethical risks.

\subsection{Limitations}
\label{sub:discussion:limitation}
Our paper has a few limitations. First, although we carefully inspect the quality of our dataset and cover different categories of LLM, the the range of datasets and the types of LLMs included are still limited. Our findings are based on the observations of these datasets and models, which may not generalize to other cases.
Related, our exploration of zero-shot few-shot prompt design is not comprehensive.  The limited input window of some models also limits our exploration of more samples for few-shot prompting.
Furthermore, we have not conducted a systematic evaluation of these models' performance in mental health reasoning.
Future work can design larger-scale experiments to include more datasets, models, prompt designs, and better evaluation.

\minorreview{Second, our datasets were mainly from Reddit, which could be limited. Although our analysis in Section~\ref{subsub:results:finetune:generalizability} shows that finetuned models have cross-platform generalizability, the finetuning was only based on Reddit and can introduce bias.} Meanwhile, although the labels are not directly accessible on the platforms, it is possible that these text data have been included in the initial training of these large models. We still argue that there is little information leakage as long as the models haven't seen the labels for our tasks, but it is hard to measure how the initial training process may affect the outcomes in our evaluation.

Third, another important limitation of the current work is the lack of evaluation of model fairness. Our anonymous datasets do not include comprehensive demographic information, making it hard to compare the performance across different population groups. As we discussed in the previous section, lots of future work on ethics and fairness is necessary before deploying such systems in real life.

\section{Conclusion}
\label{sec:conclusion}
In this paper, we present the first comprehensive evaluation of multiple LLMs (\review{Alpaca, Alpaca-LoRA, FLAN-T5, LLaMA2, GPT-3.5, and GPT-4}) on mental health prediction tasks (binary and multi-class classification) via online text data. 
Our experiments cover zero-shot prompting, few-shot prompting, and instruction finetuning. The results reveal a number of interesting findings.
Our context enhancement strategy can robustly improve performance for all LLMs, and our mental health enhancement strategy can enhance models with a large number of trainable parameters.
Meanwhile, few-shot prompting can also robustly improve model performance even by providing just one example per class.
\review{Most importantly, our experiments show that instruction finetuning across multiple datasets can significantly boost model performance on various mental health prediction tasks at the same time, generalizing across external data sources and platforms.}. Our best finetuned models, Mental-Alpaca and Mental-FLAN-T5, outperform much larger \review{LLaMA2, GPT-3.5 and GPT-4}, and perform on par with the state-of-the-art task-specific model Mental-RoBERTa.
We also conduct an exploratory case study on these models' reasoning capability, which further suggests both the promising future and the important limitations of LLMs.
We summarize our findings as a set of guidelines for future researchers, developers, and practitioners who want to empower LLMs with better knowledge of mental health for downstream tasks.
Meanwhile, we emphasize that our current efforts of LLMs in mental health are still far from deployability. We highlight the important ethical concerns accompanying this line of research.

\section*{ACKNOWLEDGMENTS}
This work is supported by VW Foundation, Quanta Computing, and the National Institutes of Health (NIH) under Grant No. 1R01MD018424-01.

\bibliographystyle{ACM-Ref-Format}
\bibliography{
bib/BehaviorIntervention,
bib/HumanComputerInteraction,
bib/Modeling_Behavior-General,
bib/MachineLearning,
bib/OrsonPublication,
bib/others
}

\newpage
\section*{Appendix: Detailed Results Tables}
\label{appendix:detail_results}

\renewcommand{\arraystretch}{1}
\begin{table}[!b]
\centering
\caption{
\review{Balanced Accuracy Performance Summary of Zero-shot, Few-shot and Instruction Finetuning on LLMs. 
$context$, $mh$, and $both$ indicate the prompt design strategies of context enhancement, mental health enhancement, and their combination  (see Table.~\ref{tab:prompt_design}).
Small numbers represent standard deviation across different designs of \textit{Prompt}$_{\textit{Part1-S}}$ and \textit{Prompt}$_{\textit{Part2-Q}}$. The baselines at the top rows do not have standard deviation as the task-specific output is static, and prompt designs do not apply.
Due to token limit, computation cost, and resource constraints, some infeasible experiments are marked as ``--''.
For each column, the best result is \textbf{bolded}, and the second best is \underline{underlined}.
}
}
\label{tab:results_overall}
\resizebox{1\textwidth}{!}{
\begin{tabular}{llccccccccc}
\thickhlinespace
& \makecell[r]{\textbf{Dataset}} & \textbf{Dreaddit} & \multicolumn{2}{c}{\textbf{DepSeverity}} & \textbf{SDCNL} & \multicolumn{2}{c}{\textbf{CSSRS-Suicide}} & \textbf{Red-Sam} & \textbf{Twt-60Users} & \textbf{SAD} \\ \addlinespace[1ex]
\textbf{Category} & \textbf{Model}   & \textbf{Task \#1}       & \textbf{Task \#2}             & \textbf{Task \#3}             & \textbf{Task \#4}    & \textbf{Task \#5}     & \textbf{Task \#6} & \textbf{Task \#2} & \textbf{Task \#2} & \textbf{Task \#1} \\ \thickhlinespace
\multirow{23}{*}{\makecell{Zero-shot\\Prompting}}  & Alpaca$_{ZS}$ & 0.593$_{\pm0.039}$ & 0.522$_{\pm0.022}$ & 0.431$_{\pm0.050}$ & 0.493$_{\pm0.007}$ & 0.518$_{\pm0.037}$ & 0.232$_{\pm0.076}$ & 0.524$_{\pm0.014}$ & 0.521$_{\pm0.022}$ & 0.503$_{\pm0.004}$\\
 & Alpaca$_{ZS-context}$ & 0.612$_{\pm0.065}$ & 0.567$_{\pm0.077}$ & 0.454$_{\pm0.143}$ & 0.497$_{\pm0.006}$ & 0.532$_{\pm0.033}$ & 0.250$_{\pm0.060}$ & 0.525$_{\pm0.019}$ & 0.559$_{\pm0.064}$ & 0.501$_{\pm0.004}$\\
 & Alpaca$_{ZS\_mh}$ & 0.593$_{\pm0.031}$ & 0.577$_{\pm0.028}$ & 0.444$_{\pm0.090}$ & 0.482$_{\pm0.015}$ & 0.523$_{\pm0.013}$ & 0.235$_{\pm0.033}$ & 0.527$_{\pm0.006}$ & 0.569$_{\pm0.017}$ & 0.522$_{\pm0.027}$\\
 & Alpaca$_{ZS\_both}$ & 0.540$_{\pm0.029}$ & 0.559$_{\pm0.040}$ & 0.421$_{\pm0.095}$ & 0.532$_{\pm0.005}$ & 0.511$_{\pm0.011}$ & 0.221$_{\pm0.030}$ & 0.495$_{\pm0.016}$ & 0.499$_{\pm0.004}$ & 0.557$_{\pm0.041}$ \\ \cdashlinespace{2-11}
& Alpaca-LoRA$_{ZS}$ & 0.571$_{\pm0.043}$ & 0.548$_{\pm0.027}$ & 0.437$_{\pm0.044}$ & 0.502$_{\pm0.011}$ & 0.540$_{\pm0.012}$ & 0.187$_{\pm0.053}$ & 0.577$_{\pm0.004}$ & 0.607$_{\pm0.046}$ & 0.477$_{\pm0.016}$\\
 & Alpaca-LoRA$_{ZS_context}$ & 0.537$_{\pm0.047}$ & 0.501$_{\pm0.001}$ & 0.343$_{\pm0.152}$ & 0.472$_{\pm0.020}$ & 0.567$_{\pm0.038}$ & 0.214$_{\pm0.059}$ & 0.535$_{\pm0.017}$ & 0.649$_{\pm0.021}$ & 0.443$_{\pm0.047}$\\
 & Alpaca-LoRA$_{ZS\_mh}$ & 0.500$_{\pm0.000}$ & 0.500$_{\pm0.000}$ & 0.331$_{\pm0.145}$ & 0.497$_{\pm0.025}$ & 0.557$_{\pm0.023}$ & 0.216$_{\pm0.022}$ & 0.541$_{\pm0.016}$ & 0.569$_{\pm0.019}$ & 0.471$_{\pm0.033}$\\
 & Alpaca-LoRA$_{ZS\_both}$ & 0.500$_{\pm0.000}$ & 0.500$_{\pm0.000}$ & 0.386$_{\pm0.059}$ & 0.499$_{\pm0.023}$ & 0.517$_{\pm0.031}$ & 0.224$_{\pm0.049}$ & 0.507$_{\pm0.009}$ & 0.535$_{\pm0.025}$ & 0.420$_{\pm0.019}$\\ \cdashlinespace{2-11}
 & FLAN-T5$_{ZS}$  & 0.659$_{\pm0.086}$ & 0.664$_{\pm0.011}$ & 0.396$_{\pm0.006}$ & 0.643$_{\pm0.021}$ & 0.667$_{\pm0.023}$ & 0.418$_{\pm0.012}$ & 0.554$_{\pm0.034}$ & 0.613$_{\pm0.040}$ & 0.692$_{\pm0.093}$\\
 & FLAN-T5$_{ZS_context}$ & 0.663$_{\pm0.079}$ & 0.674$_{\pm0.014}$ & 0.378$_{\pm0.013}$ & 0.653$_{\pm0.011}$ & 0.649$_{\pm0.026}$ & 0.378$_{\pm0.029}$ & 0.563$_{\pm0.029}$ & 0.613$_{\pm0.046}$ & 0.738$_{\pm0.056}$\\
 & FLAN-T5$_{ZS\_mh}$ & 0.616$_{\pm0.070}$ & 0.666$_{\pm0.009}$ & 0.366$_{\pm0.012}$ & 0.648$_{\pm0.010}$ & 0.653$_{\pm0.018}$ & 0.372$_{\pm0.033}$ & 0.547$_{\pm0.035}$ & 0.613$_{\pm0.033}$ & 0.739$_{\pm0.039}$\\
 & FLAN-T5$_{ZS\_both}$ & 0.604$_{\pm0.074}$ & 0.661$_{\pm0.004}$ & 0.389$_{\pm0.051}$ & 0.645$_{\pm0.005}$ & 0.657$_{\pm0.019}$ & 0.382$_{\pm0.048}$ & 0.536$_{\pm0.027}$ & 0.606$_{\pm0.040}$ & 0.767$_{\pm0.050}$\\ \cdashlinespace{2-11}
 & LLaMA2$_{ZS}$ & 0.720$_{\pm0.012}$ & 0.693$_{\pm0.034}$ & 0.429$_{\pm0.013}$ & 0.589$_{\pm0.010}$ & 0.691$_{\pm0.014}$ & 0.261$_{\pm0.018}$ & 0.574$_{\pm0.008}$ & 0.735$_{\pm0.017}$ & 0.704$_{\pm0.026}$\\
 & LLaMA2$_{ZS_context}$ & 0.658$_{\pm0.025}$ & 0.707$_{\pm0.056}$ & 0.410$_{\pm0.019}$ & 0.588$_{\pm0.026}$ & 0.722$_{\pm0.039}$ & 0.367$_{\pm0.043}$ & 0.562$_{\pm0.011}$ & \underline{0.736}$_{\pm0.019}$ & 0.650$_{\pm0.027}$\\
 & LLaMA2$_{ZS\_mh}$ & 0.617$_{\pm0.012}$ & 0.711$_{\pm0.033}$ & 0.395$_{\pm0.017}$ & 0.642$_{\pm0.008}$ & 0.696$_{\pm0.021}$ & 0.291$_{\pm0.038}$ & 0.572$_{\pm0.012}$ & 0.689$_{\pm0.056}$ & 0.567$_{\pm0.021}$\\
 & LLaMA2$_{ZS\_both}$ & 0.584$_{\pm0.017}$ & 0.704$_{\pm0.036}$ & 0.444$_{\pm0.021}$ & 0.643$_{\pm0.014}$ & 0.689$_{\pm0.043}$ & 0.328$_{\pm0.058}$ & 0.559$_{\pm0.012}$ & 0.692$_{\pm0.069}$ & 0.560$_{\pm0.009}$\\ \cdashlinespace{2-11}
 & GPT-3.5$_{ZS}$ & 0.685$_{\pm0.024}$ & 0.642$_{\pm0.017}$ & 0.603$_{\pm0.017}$ & 0.460$_{\pm0.163}$ & 0.570$_{\pm0.118}$ & 0.233$_{\pm0.009}$ & 0.454$_{\pm0.007}$ & 0.536$_{\pm0.024}$ & 0.717$_{\pm0.017}$\\
 & GPT-3.5$_{ZS_context}$ & 0.688$_{\pm0.045}$ & 0.653$_{\pm0.020}$ & 0.543$_{\pm0.047}$ & 0.618$_{\pm0.008}$ & 0.577$_{\pm0.090}$ & 0.265$_{\pm0.048}$ & 0.473$_{\pm0.001}$ & 0.560$_{\pm0.002}$ & 0.723$_{\pm0.003}$\\
 & GPT-3.5$_{ZS\_mh}$ & 0.679$_{\pm0.017}$ & 0.636$_{\pm0.021}$ & 0.642$_{\pm0.034}$ & 0.576$_{\pm0.001}$ & 0.477$_{\pm0.014}$ & 0.310$_{\pm0.015}$ & 0.467$_{\pm0.004}$ & 0.571$_{\pm0.000}$ & 0.664$_{\pm0.061}$\\
 & GPT-3.5$_{ZS\_both}$ & 0.681$_{\pm0.010}$ & 0.627$_{\pm0.022}$ & 0.617$_{\pm0.014}$ & 0.632$_{\pm0.020}$ & 0.617$_{\pm0.033}$ & 0.254$_{\pm0.009}$ & 0.506$_{\pm0.004}$ & 0.570$_{\pm0.007}$ & 0.750$_{\pm0.027}$\\ \cdashlinespace{2-11}
 & GPT-4$_{ZS}$ & 0.700$_{\pm0.001}$ & 0.719$_{\pm0.013}$ & 0.588$_{\pm0.010}$ & 0.644$_{\pm0.007}$ & 0.760$_{\pm0.009}$ & 0.418$_{\pm0.009}$ & 0.434$_{\pm0.005}$ & 0.566$_{\pm0.017}$ & \textbf{0.854}$_{\pm0.006}$\\
 & GPT-4$_{ZS_context}$ & 0.706$_{\pm0.009}$ & 0.719$_{\pm0.009}$ & 0.590$_{\pm0.011}$ & 0.644$_{\pm0.011}$ & 0.753$_{\pm0.028}$ & \underline{0.441}$_{\pm0.057}$ & 0.465$_{\pm0.010}$ & 0.565$_{\pm0.006}$ & \underline{0.848}$_{\pm0.001}$\\
 & GPT-4$_{ZS\_mh}$ & 0.725$_{\pm0.009}$ & 0.684$_{\pm0.004}$ & 0.656$_{\pm0.001}$ & 0.645$_{\pm0.012}$ & 0.737$_{\pm0.005}$ & 0.396$_{\pm0.020}$ & 0.496$_{\pm0.005}$ & 0.527$_{\pm0.007}$ & 0.840$_{\pm0.003}$\\
 & GPT-4$_{ZS\_both}$ & 0.719$_{\pm0.021}$ & 0.689$_{\pm0.000}$ & 0.650$_{\pm0.011}$ & 0.647$_{\pm0.014}$ & 0.697$_{\pm0.005}$ & 0.411$_{\pm0.009}$ & 0.511$_{\pm0.000}$ & 0.546$_{\pm0.014}$ & 0.837$_{\pm0.002}$\\
\thickhlinespace
\multirow{4}{*}{\makecell{Few-shot\\Prompting}} &  Alpaca$_{FS}$         & 0.632$_{\pm0.030}$ & 0.529$_{\pm0.017}$ & 0.628$_{\pm0.005}$ & ---   & ---                & ---                & ---             & ---                & ---                \\
& FLAN-T5$_{FS}$         & 0.786$_{\pm0.006}$ & 0.678$_{\pm0.009}$ & 0.432$_{\pm0.009}$ & ---                & ---                & ---   & ---                & ---                & ---             \\
& GPT-3.5$_{FS}$         & 0.721$_{\pm0.010}$ & 0.665$_{\pm0.015}$ & 0.580$_{\pm0.002}$ & ---                & ---                & --- & ---                & ---                & ---               \\
& GPT-4$_{FS}$         & 0.698$_{\pm0.009}$ & 0.724$_{\pm0.005}$ & 0.613$_{\pm0.001}$ & ---                & ---                & --- & ---                & ---                & ---                \\\thickhlinespace
\multirow{2}{*}{\makecell{Instructional\\Finetuning}} & Mental-Alpaca         & \underline{0.816}$_{\pm0.006}$ & \underline{0.775}$_{\pm0.006}$ & \underline{0.746}$_{\pm0.005}$ & \textbf{0.724}$_{\pm0.004}$ & 0.730$_{\pm0.048}$ & 0.403$_{\pm0.029}$ & \textbf{0.604}$_{\pm0.012}$ & 0.718$_{\pm0.011}$ & 0.819$_{\pm0.006}$\\
& Mental-FLAN-T5 &  0.802$_{\pm0.002}$ & 0.759$_{\pm0.003}$ & \textbf{0.756}$_{\pm0.001}$ & 0.677$_{\pm0.005}$ & \textbf{0.868}$_{\pm0.006}$ & \textbf{0.481}$_{\pm0.006}$ & \underline{0.582}$_{\pm0.002}$ & \textbf{0.736}$_{\pm0.003}$ & 0.779$_{\pm0.002}$ \\ \thickhlinespace
\multirow{3}{*}{\xspace\xspace\xspace Baseline} & Majority          & 0.500$_{\pm ---}$  & 0.500$_{\pm ---}$  & 0.250$_{\pm ---}$  & 0.500$_{\pm ---}$  & 0.500$_{\pm ---}$  & 0.200$_{\pm ---}$ & ---                & ---                & --- \\
& BERT              & 0.783$_{\pm ---}$  & 0.763$_{\pm ---}$  & 0.690$_{\pm ---}$  & 0.678$_{\pm ---}$  & 0.500$_{\pm ---}$  & 0.332$_{\pm ---}$ & ---                & ---                & --- \\
& Mental-RoBERTa     & \textbf{0.831}$_{\pm ---}$  & \textbf{0.790}$_{\pm ---}$   & 0.736$_{\pm ---}$  & \underline{0.723}$_{\pm ---}$  & \underline{0.853}$_{\pm ---}$  & 0.373$_{\pm ---}$ & ---                & ---                & --- \\ \thickhlinespace
\end{tabular}
\vspace{-0.5cm}
}
\end{table}
\renewcommand{\arraystretch}{1}

\end{document}